\documentclass{article}

\usepackage{microtype}
\usepackage{graphicx}
\usepackage{caption}
\usepackage{subcaption}
\usepackage{tikz}
\usetikzlibrary{bayesnet}
\usepackage{booktabs}
\usepackage{amsfonts}
\usepackage{amsmath}
\usepackage{nccmath}
\usepackage{enumitem}
\usepackage{amsthm}
\usepackage{dsfont}
\usepackage{lipsum}

\usepackage{hyperref}

\usepackage[accepted]{icml2022}

\icmltitlerunning{Bayesian Nonparametrics for Offline Skill Discovery}

\begin{document}

\twocolumn[
\icmltitle{Bayesian Nonparametrics for Offline Skill Discovery}



\icmlsetsymbol{equal}{*}

\begin{icmlauthorlist}
\icmlauthor{Valentin Villecroze}{L6,to}
\icmlauthor{Harry J. Braviner}{L6}
\icmlauthor{Panteha Naderian}{L6}
\icmlauthor{Chris J. Maddison}{to,vec}
\icmlauthor{Gabriel Loaiza-Ganem}{L6}
\end{icmlauthorlist}

\icmlaffiliation{L6}{Layer 6 AI, Toronto, Canada}
\icmlaffiliation{to}{University of Toronto, Toronto, Canada}
\icmlaffiliation{vec}{Vector Institute, Toronto, Canada}

\icmlcorrespondingauthor{Valentin Villecroze}{valentin.v@layer6.ai}
\icmlcorrespondingauthor{Harry J. Braviner}{harry@layer6.ai}
\icmlcorrespondingauthor{Panteha Naderian}{panteha@layer6.ai}
\icmlcorrespondingauthor{Chris J. Maddison}{cmaddis@cs.toronto.edu}
\icmlcorrespondingauthor{Gabriel Loaiza-Ganem}{gabriel@layer6.ai}

\icmlkeywords{Imitation learning, Bayesian nonparametrics, Reinforcement learning, Variational inference}

\vskip 0.3in
]



\printAffiliationsAndNotice{}  

\begin{abstract}
Skills or low-level policies in reinforcement learning are temporally extended actions that can speed up learning and enable complex behaviours. Recent work in offline reinforcement learning and imitation learning has proposed several techniques for skill discovery from a set of expert trajectories. While these methods are promising, the number $K$ of skills to discover is always a fixed hyperparameter, which requires either prior knowledge about the environment or an additional parameter search to tune it. We first propose a method for offline learning of options (a particular skill framework) exploiting advances in variational inference and continuous relaxations. We then highlight an unexplored connection between Bayesian nonparametrics and offline skill discovery, and show how to obtain a nonparametric version of our model. This version is tractable thanks to a carefully structured approximate posterior with a dynamically-changing number of options, removing the need to specify $K$. We also show how our nonparametric extension can be applied in other skill frameworks, and empirically demonstrate that our method can outperform state-of-the-art offline skill learning algorithms across a variety of environments. Our code is available at \url{https://github.com/layer6ai-labs/BNPO}.
\end{abstract}

\section{Introduction}

Hierarchical policies have been studied in reinforcement learning for decades \citep{dietterich1998maxq, vezhnevets2017feudal}. Low-level policies in hierarchical frameworks can be interpreted as skills endowing agents with temporal abstraction. These skills also offer a divide-and-conquer approach to complex reinforcement learning environments, and learning them is therefore a relevant problem.

One of the most commonly used skills frameworks is options \citep{sutton1999between}. Methods for online learning of options have been proposed \citep{bacon2017option, khetarpal2020options}, but cannot leverage data from expert trajectories. To the best of our knowledge, the only method allowing offline option discovery is that of \citet{fox2017multi} (\textit{deep discovery of options}, DDO), which forgoes the use of highly successful variational inference advances \citep{kingma2013auto, rezende2014stochastic} for discrete latent variables \citep{maddison2016concrete, jang2016categorical} which in this case correspond to the options to be inferred. Our first contribution is proposing a method for offline learning of options combining these previously neglected advances along with a judiciously constructed approximate posterior, which we show empirically outperforms not only DDO, but also other offline skill discovery algorithms.

Additionally, all (discrete) skill learning approaches we are aware of -- including options -- require specifying the number of skills $K$ as a hyperparameter, rather than learning it. This is a significant practical limitation, as $K$ cannot realistically be known in advance for complex environments. Our second contributions is highlighting a similarity between mixture models and inferring skills from expert trajectories: both are a form of soft clustering where observed data (state-action pairs) are probabilistically assigned to their corresponding cluster (skill). Bayesian nonparametric approaches such as Dirichlet process mixture models \citep{neal2000markov} are not only mathematically elegant, but also circumvent the practical need to prespecify the number of clusters. We therefore argue that Bayesian nonparametrics should be more heavily used for option and skill discovery.

Our third contribution is proposing a scheme allowing for our option discovery method to accommodate a nonparametric prior over options (thus modelling $K$ as infinite), which can also be applied in other skill learning frameworks. By further adding structure to the variational posterior, allowing it to dynamically change the number of options/skills being assigned positive probability, we recover a method that is not only tractable, but also retains the nonparametric aspect of the model and does not require treating $K$ as a hyperparameter. We show that our nonparametric versions of skill learning algorithms match the performance of their parametric counterparts with tuned $K$, thus successfully ``learning a sufficient number of options'' in practice. Finally, we hope that our nonparametric variational inference scheme will find uses outside of offline skill recovery.

\section{Related Work}

Imitation learning \citep{hussein2017imitation}, also called \textit{learning from demonstrations} \citep{schaal1997learning, argall2009survey}, consists of learning to solve a task from a set of expert demonstrations. This can be achieved by methods such as behavioural cloning \citep{esmaili1995behavioural, ross2011reduction} or inverse reinforcement learning \citep{ho2016generative, sharma2018directed, krishnan2019swirl}. In this paper we focus on the former. Many recent works on imitation learning attempt to divide the demonstrations into several smaller segments, before fitting models on each of them \citep{niekum2013incremental, murali2016tsc, krishnan2018transition, shiarlis2018taco}. These works usually consider these two steps as two distinct stages, and unlike our proposed approach, do not learn skills end-to-end. \citet{niekum2013incremental} and \citet{krishnan2018transition} are of particular interest as they are the only ones, to our knowledge, to make use of a Bayesian nonparametric model to segment the trajectories. However, they significantly differ from our method in that they use handcrafted movement primitives and linear dynamical models to fit the resulting segments.

Our work is closer to the field of option and skill discovery, which leverages work on online hierarchical reinforcement learning \citep{sutton1999between, gregor2016variational, bacon2017option, achiam2018variational, eysenbach2018diversity, sharma2019dynamics, florensa2017stochastic} and aims to learn adequate options from a set of expert trajectories. DDO \citep{fox2017multi} uses an EM-type algorithm \citep{dempster1977maximum} that allows multi-level hierarchies. As previously mentioned, this method forgoes advances in variational inference and continuous relaxations, and we will later show that option learning can be significantly improved upon through the use of these techniques. While not following the option framework, \citet{kipf2019compile} propose CompILE, which uses a variational approach and continuous relaxations to both segment the trajectories and encode the resulting slices as discrete skills. We will show that with these advances, options also outperform CompILE.

Both DDO and CompILE need the number of options/skills, $K$, to be specified beforehand. We borrow from the Bayesian nonparametrics literature \citep{teh2006hierarchical, caron2007generalized, dunson2009nonparametric} -- in particular from models involving Dirichlet processes \citep{ferguson1973bayesian} -- in order to model options. We follow \citet{nalisnick2016stick} and use a variational-autoencoder-inspired \citep{kingma2013auto, rezende2014stochastic} inference scheme that avoids the need for the expensive MCMC samplers commonly associated with these types of models \citep{neal2000markov}. As a result, the nonparametric versions of the models we consider do not need to prescpecify nor tune $K$.

Finally, \citet{shankar2020learning} and \citet{ajay2021opal} also use variational inference to discover skills but choose to represent them as continuous latent variables instead of categorical variables. While this choice also technically leads to infinitely many skills, as we will see in our experiments, discrete skills learned from expert trajectories are a useful way to enhance online agents. The use of continuous skills results not only in less interpretable skills, but also loses the ability to perform these enhancements.
\section{Preliminaries}

We now review the main concepts needed for our model: behavioural cloning, the options framework, continuous relaxations, and the stick-breaking process. We also review CompILE, a method that we shall later make nonparametric.

\subsection{Behavioural Cloning}
The goal of behavioural cloning is to imitate an expert who demonstrates how to accomplish a task.
More formally, we consider a Markov Decision Process without rewards (MDP\textbackslash R). An MDP\textbackslash R is a tuple $\mathcal{M}:\langle \mathcal{S}, \mathcal{A}, P, \rho\rangle$, where $\mathcal{S}$ is the state space, $\mathcal{A}$ the action space, $P\left(s_{t+1} \mid s_{t}, a_{t}\right)$ a transition distribution, and $\rho$ the starting state distribution. We assume access to a dataset of expert trajectories $ \xi := \{\xi^{(i)}:=(s_0^{(i)}, a_0^{(i)}, s_1^{(i)}, a_1^{(i)}, \dots, s_T^{(i)})\}^N_{i=1}$ of states $s_t^{(i)} \in \mathcal{S}$ and actions $a_t^{(i)} \in \mathcal{A}$.
The trajectory length $T$ need not be identical across trajectories and could be changed to $T_i$, but we keep $T$ to avoid further complicating notation. We want to find a policy $\pi_\theta: \mathcal{S} \mapsto \Delta(\mathcal{A})$ (where $\Delta(\mathcal{A})$ denotes the set of distributions over $\mathcal{A}$) parameterized by $\theta$ that maximizes the logarithm of following expression:
\begin{align}
    p_\theta (\xi)
& = \medmath{\prod_{i=1}^N \rho(s^{(i)}_0)
\prod_{t=0}^{T-1}
\pi_\theta ( a_t^{(i)} | s_t^{(i)}) P(s_{t+1}^{(i)} | s_t^{(i)}, a_t^{(i)})} \label{eqn:bc1}\\
& \propto \prod_{i=1}^N \prod_{t=0}^{T-1}
\pi_\theta ( a_t^{(i)} | s_t^{(i)}), \label{eqn:bc2}
\end{align}
i.e. the policy maximizing the likelihood assigned to the expert trajectories. We have dropped terms that are constant with respect to $\theta$ in Equation \ref{eqn:bc2}.

\subsection{The Options Framework}
As previously mentioned, the option framework introduces temporally extended actions, allowing for temporal abstraction and more complex behaviour. Formally, an option $h \in \Omega$ \citep{sutton1999between} is a tuple $\langle\mathcal{I}_{h}, \pi_{h}, \psi_{h}\rangle$, where  $\mathcal{I}_{h} \subseteq \mathcal{S}$ is the initiation set, $\pi_{h}(a|s)$ the control or low-level policy, and $\psi_{h}:\mathcal{S} \rightarrow [0,1]$ the termination function. An option $h$ can be invoked in any state $s \in \mathcal{I}_{h}$, in which case actions are drawn according to $\pi_h$ until the option terminates, which happens with probability $\psi_h(s')$ at each subsequent state $s'$. In our setting we assume that $\mathcal{I}_{h} = \mathcal{S}$, which is a common assumption \citep{bacon2017option, fox2017multi, shankar2020learning}. 
In order to use these options to solve a task, a high-level policy (or policy over options) $\eta : \mathcal{S} \mapsto \Delta(\Omega)$ is used to select a new option $h$ after the previous one terminates. This two-level structure is called a hierarchical policy. The resulting generative process is described in Algorithm \ref{alg:algo0}.

Learning a hierarchical policy from expert demonstrations using behavioural cloning is much more challenging than with a single policy. Indeed, not only do multiple policies need to be learned concurrently (one for each option), but, as the options are unobserved, they must be treated as latent variables. \citet{fox2017multi} propose an EM-based approach to do this, which we improve upon.

\begin{algorithm}[t]
   \caption{Trajectory generation with options.}
   \label{alg:algo0}
    \begin{algorithmic}[1]
       \STATE $s_0 \sim \rho(\cdot)$, $b_0 \leftarrow 1$ 
       \FOR{$t \in [0, \dots, T-1]$}
       \IF{$b_t = 1$}
       \STATE{$h_t \sim \eta(\cdot|s_t)$} \textit{\small{\textbackslash\textbackslash Draw option from high-level policy}}
       \ELSE
       \STATE{$h_t \leftarrow h_{t-1}$}
       \ENDIF
       \STATE $a_t \sim \pi_{h_t}(\cdot| s_t)$ \textit{\small{\textbackslash\textbackslash Draw action from low-level policy}}
       \STATE $s_{t+1} \sim P(\cdot|s_t, a_t)$ \textit{\small{\textbackslash\textbackslash Draw the next state}}
       \STATE $b_{t+1} \sim \operatorname{Bernoulli}(\cdot | \psi_{h_{t}}(s_{t+1}))$ \textit{\small{\textbackslash\textbackslash Terminate option?}}
       \ENDFOR
    \end{algorithmic}
\end{algorithm}

\subsection{CompILE}

CompILE \citep{kipf2019compile} is used for hierarchical imitation learning but does not rely on the option framework. Binary termination variables are not sampled at every timestep to determine if the current skill should terminate. Rather, whenever a new skill is drawn from the high-level policy, CompILE also samples the number of steps for which the skill will be active from a Poisson distribution with parameter $\lambda$.
Also, rather than using a state-dependent high-level policy, $\eta$ is assumed to be uniform over skills.
We refer to any time interval between subsequent skill selections as a segment.
This procedure is summarized in Algorithm \ref{alg:algo1}, where we have abused notation and still use $b$ to denote termination variables in order to highlight the similarity with the options framework, even though these are not the same variables as in options. Similarly, $h_j$ does not denote an option, but rather the skill being used during segment $j$.

Doing behavioural cloning in ComPILE is harder than just maximizing Equation \ref{eqn:bc2}, since computing $\log p_\theta(\xi)$, where $\theta$ now parameterizes all the low-level policies, requires an intractable marginalization over the unobserved variables ($b$'s and $h$'s). In order to circumvent this issue, an approximate posterior $q_\phi(\zeta|\xi)$ is introduced, where $\zeta$ denotes all the unobserved variables for all trajectories and $\phi$ the variational parameters. Instead of directly maximizing $\log p_\theta(\xi)$, the following lower bound, which is called the ELBO, is maximized over $(\theta, \phi)$:
\begin{align}\label{eq:comp}
L(\theta, \phi) & := \mathbb{E}_{q_\phi(\zeta|\xi)}[\log p_\theta(\zeta, \xi) - \log q_\phi(\zeta|\xi)]\\
& \leq \log p_\theta(\xi).
\end{align}
We omit the details on how $q_\phi(\zeta|\xi)$ is structured, but highlight that the number of segments needs to be specified as a hyperparameter of this variational approximation rather than being properly treated as random. In order to maximize Equation \ref{eq:comp}, the authors use the reparameterization trick of variational autoencoders \citep{kingma2013auto,rezende2014stochastic} along with continuous relaxations. In particular, they use the Concrete or Gumbel-Softmax (GS) distribution \citep{maddison2016concrete, jang2016categorical}, which we briefly review in the next section, in order to efficiently backpropagate through Equation \ref{eq:comp}.

\begin{algorithm}[t]
   \caption{Trajectory generation with CompILE.}
   \label{alg:algo1}
    \begin{algorithmic}[1]
       \STATE $s_0 \sim \rho(\cdot)$, $b_0 \leftarrow 0$, $j \leftarrow 0$ \\
       \FOR{$t \in [0, \dots, T-1]$}
       \WHILE{$t = b_j$}
       \STATE $j \leftarrow j+1$ \textit{\small{\textbackslash\textbackslash Consider the next segment}}
        \STATE{$h_j \sim \eta(\cdot) = \operatorname{Uniform}(\cdot | K)$} \textit{\small{\textbackslash\textbackslash Draw skill}}
       \STATE $b_j \sim \operatorname{Poisson}(\cdot | \lambda)$
       \textit{\small{\textbackslash\textbackslash Draw segment length}}
       \STATE $b_j \leftarrow b_j + b_{j-1}$
       \textit{\small{\textbackslash\textbackslash Next segment boundary}}
       \ENDWHILE
       \STATE $a_t \sim \pi_{h_j}(\cdot| s_t)$ \textit{\small{\textbackslash\textbackslash Draw action from low-level policy}}
       \STATE $s_{t+1} \sim P(\cdot|s_t, a_t)$ \textit{\small{\textbackslash\textbackslash Draw the next state}}
       \ENDFOR
    \end{algorithmic}
\end{algorithm}

\subsection{Continuous Relaxations}

The reparameterization trick is used in variational autoencoders to backpropagate through expressions of the form $\mathbb{E}_{q_\phi(\zeta)}[f_\phi(\zeta)]$,
with respect to $\phi$, where $f_\phi$ is a real-valued function. Since the distribution with respect to which the expectation is taken depends on $\phi$, one cannot simply bring the gradient inside of the expectation. Gradient estimators such as REINFORCE \citep{glynn1990likelihood, williams1992simple} typically exhibit high variance, which the reparameterization trick empirically reduces. When $\zeta$ is a continuous random variable, it is often (but not always) the case that one can easily find a continuously differentiable function $g$ such that $\zeta \sim q_\phi(\cdot) \iff \zeta=g(\epsilon, \phi)$ where $\epsilon$ follows some continuous distribution which does not depend on $\phi$. In this case, the gradient is given by:
\begin{equation}\label{eq:reparam}
\nabla_\phi \mathbb{E}_{q_\phi(\zeta)}[f_\phi(\zeta)] = \mathbb{E}_{\epsilon}[\nabla_\phi f_\phi(g(\epsilon, \phi))],
\end{equation}
and a Monte Carlo estimate can be easily obtained.

When $\zeta$ is categorical, $g$ has to be piece-wise constant and Equation \ref{eq:reparam} no longer holds. Continuous relaxations approximate $\mathbb{E}_{q_\phi(\zeta)}[f_\phi(\zeta)]$ with an expectation over a continuous random variable, so that the reparameterization trick can be used.
The Gumbel-Softmax (GS) or Concrete distribution is a distribution on the $K$-simplex, $\Delta(K):=\{x\in \mathbb{R}^K: x_k > 0, \sum_{k=0}^{K-1} x_k=1\}$, parameterized by $q \in \Delta(K)$ and a temperature hyperparameter $\tau > 0$, designed to address this issue. It is reparameterized as follows:
\begin{equation}
    \zeta \sim GS_\tau(\cdot|q) \iff \zeta = \operatorname{softmax}\left(\dfrac{\epsilon + \log q}{\tau}\right),
\end{equation}
where $\epsilon$ is a $K$-dimensional vector with independent Gumbel$(0,1)$ entries, and the $\log$ is taken elementwise. As $\tau \rightarrow 0$, the $GS_\tau(\cdot|q)$ distribution converges to the discrete distribution $q$. By thinking of categorical $\zeta$'s as one-hot vectors of length $K$, the GS thus provides a continuous relaxation of $\zeta$, and $\mathbb{E}_{q_\phi(\zeta)}[f_\phi(\zeta)]$ can be approximated by:
\begin{equation}\label{eq:gs_reparam}
    \mathbb{E}_{q_\phi(\zeta)}[f_\phi(\zeta)] \approx \mathbb{E}_{GS_\tau(\zeta | q_\phi(\cdot))}[\tilde{f}(\zeta)],
\end{equation}
where we think of the discrete distribution $q_\phi(\cdot)$ as a $K$-dimensional vector, and $\tilde{f}$ is a relaxation of $f$ mapping all of $\Delta(K)$ (rather than just the vertices, i.e. one-hot vectors) to $\mathbb{R}$. Equation \ref{eq:gs_reparam} admits the gradient estimator of Equation \ref{eq:reparam}, and \citet{kipf2019compile} use it to optimize Equation \ref{eq:comp}.

\subsection{The Stick-Breaking Process}

The stick-breaking process \citep{sethuraman1994constructive, ishwaran2001gibbs}, which is deeply connected to Dirichlet processes, places a distribution over probability vectors of infinite length. Equivalently, it is a distribution over distributions on a countably infinite set. This process will allow us to both assume that there are infinitely many options, and to place a proper prior on the high-level policy $\eta$ itself (i.e. a distribution over distributions of options). When performing posterior inference some options will have extremely small probabilities; this enables us to ``learn the number of options'' from the data. More formally, the Griffiths-Engen-McCloskey distribution, denoted $\operatorname{GEM}(\alpha)$ and parameterized by $\alpha > 0$, produces an infinitely long vector $(\beta_0, \beta_1, \dots)$ such that $\beta_k > 0$ and $\sum_{k=0}^\infty \beta_k = 1$ almost surely. Samples are obtained by first sampling $\beta'_k \sim \operatorname{Beta}(1, \alpha)$ for $k=0,1,\dots$ and then setting:
\begin{equation}
    \beta_k = \beta_k' \prod_{l=0}^{k-1} (1 - \beta_l').
\end{equation}
Intuitively, we start off with a stick of length $1$, and at every step we remove $\beta_k$ from the stick. $\beta'_k$ is the fraction of the remaining stick that is assigned to $\beta_k$.

We shall later use the $\operatorname{GEM}$ distribution as a prior, and will want to perform variational inference. Applying the stick-breaking procedure to Beta random variables with different parameters might seem like the most natural approximate posterior. However, the Beta distribution is not easily reparameterized, and thus inference with such a posterior becomes computationally challenging. It is therefore common to instead use the Kumaraswamy distribution in the approximate posterior \citep{nalisnick2016stick, stirn2019new}. Like the Beta distribution, this is a $(0,1)$-valued distribution with two parameters $a_1, a_2 > 0$, whose density is given by:
\begin{equation}
    p(x|a_1,a_2) = a_1 a_2 x^{a_1-1}(1-x^{a_1})^{a_2-1},
\end{equation}
and which can easily be reparameterized.

\section{Our Model}

We first describe a parametric version of our model, where the number of options $K$ is a fixed hyperparameter. We will detail in Sections \ref{sec:gem_prior} and \ref{sec:heuristic} how we use a nonparametric prior to circumvent the need to specify this number. Throughout this section and until Section \ref{sec:npcomp} (exclusive), our notation refers to options and not to CompILE.

\subsection{Overview}

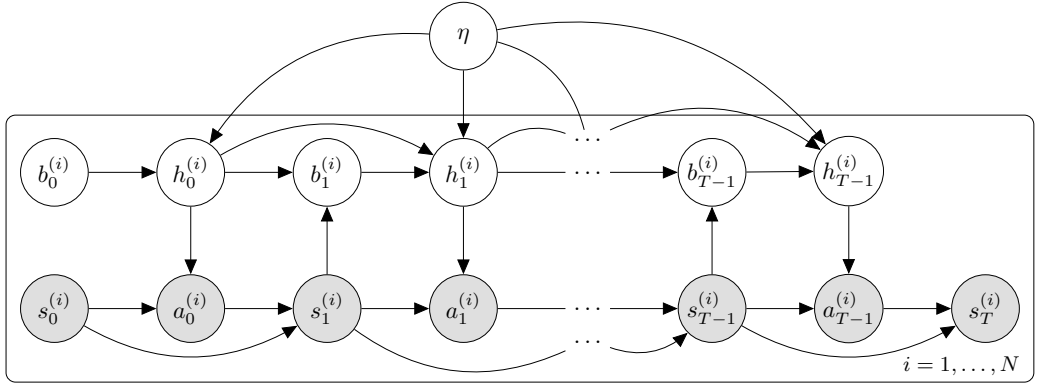
\begin{figure*}[hbtp]
    \centering
    \scalebox{0.9}{
    \begin{tikzpicture}[->, latent/.append style={minimum size=1cm},obs/.append style={minimum size=1cm}]
        \node[obs] (s0) {$s_0^{(i)}$} ;
        \node[obs, right=of s0] (a0) {$a_0^{(i)}$} ;
        \node[latent, above=of s0] (b0) {$b_0^{(i)}$} ;
        \node[latent, above=of a0] (h0) {$h_0^{(i)}$} ;
        \edge {s0} {a0} ;
        \edge {h0} {a0} ;
        \edge {b0} {h0} ;

        \node[obs, right=of a0] (s1) {$s_1^{(i)}$} ;
        \node[obs, right=of s1] (a1) {$a_1^{(i)}$} ;
        \node[latent, above=of s1] (b1) {$b_1^{(i)}$} ;
        \node[latent, above=of a1] (h1) {$h_1^{(i)}$} ;
        \edge {s1} {a1} ;
        \edge {s1} {b1} ;
        \edge {h1} {a1} ;
        \edge {b1} {h1} ;
        \edge {a0} {s1} ;
        \edge {h0} {b1} ;
        \path (s0) edge[bend right]  (s1);
        \path (h0) edge[bend left]  (h1);
        
        \node[right=of a1] (temp0) {$\ldots$} ;
        \node[below=0.2cm of temp0] (temp1) {$\ldots$} ;
        \node[right=of h1] (temp2) {$\ldots$} ;
        \node[above=0.2cm of temp2] (temp3) {$\ldots$} ;
        
        \node[obs, right=of temp0] (s3) {$s_{T-1}^{(i)}$} ;
        \node[obs, right=of s3] (a3) {$a_{T-1}^{(i)}$} ;
        \node[latent, above=of s3] (b3) {$b_{T-1}^{(i)}$} ;
        \node[latent, above=of a3] (h3) {$h_{T-1}^{(i)}$} ;
        \edge {s3} {a3} ;
        \edge {s3} {b3} ;
        \edge {h3} {a3} ;
        \edge {b3} {h3} ;

        \edge[-] {a1} {temp0} ;
        \edge[-] {h1} {temp2} ;
        \edge {temp0} {s3} ;
        \edge {temp2} {b3} ;
        \path[-] (s1) edge[bend right]  (temp1);
        \path[-] (h1) edge[bend left]  (temp3);
        \path (temp1) edge[bend right]  (s3);
        \path (temp3) edge[bend left]  (h3);
        
        \node[obs, right=of a3] (s4) {$s_{T}^{(i)}$} ;
        \edge {a3} {s4} ;
        \path (s3) edge[bend right]  (s4);
        
        \plate [inner sep=.2cm,yshift=.1cm] {plate1} {(s0)(a0)(b0)(h0)(s1)(a1)(b1)(h1)(s3)(a3)(b3)(h3)(s4)} {$i=1,\dots,N$};
        
        \node[latent, above=of h1] (eta) {$\eta$} ;
        \path (eta) edge[bend right] (h0) ;
        \edge {eta} {h1} ;
        \path[-] (eta) edge[bend left] (temp3) ;
        \path (eta) edge[bend left] (h3) ;
        
    \end{tikzpicture}
    }
    \caption{Graphical model for options. Shaded nodes correspond to observed variables ($\xi^{(i)}$) while the blank ones are latent ($\zeta^{(i)}$).}
    \label{fig:graphical_model}
\end{figure*}

Now that we have covered the required concepts, we introduce our method in more detail. We assume that the expert trajectories $\xi$ are generated by a two-level option hierarchy, as described in Algorithm \ref{alg:algo0}, with the high-level policy $\eta$ being shared across trajectories. We denote the corresponding trajectories of hidden options $h$ and binary termination variables $b$ as $\zeta := \{\zeta^{(i)}:=(b_0^{(i)}, h_0^{(i)}, b_1^{(i)}, h_1^{(i)}, \dots, h_{T-1}^{(i)})\}^N_{i=1}$, and $\eta$ the high-level policy. \citet{fox2017multi} and \citet{kipf2019compile} found that taking $\eta$ as a uniform policy rather than learning it results in equally useful learned skills. We simplify $\eta$ with the less restrictive assumption that it does not depend on the current state $s$, a choice that will later simplify our nonparametric component.
That is, in Algorithm \ref{alg:algo0}, we draw $h_t$ according to $\eta(\cdot)$ rather than $\eta(\cdot|s_t)$. We then treat $\eta$ in a Bayesian way, as a latent variable to be inferred from observed trajectories, and assume a $K$-dimensional prior over it, obtained by truncating the stick-breaking process of  Beta$(1,\alpha)$ variables after $K-1$ steps (and having the last entry be such that the resulting vector adds up to $1$).
The resulting graphical model can be seen in Figure \ref{fig:graphical_model}.

We denote as $\theta$ all the parameters from the prior (i.e. $\alpha$), the low-level policies, and the terminations functions. We implement the termination functions with a single neural network taking $s$ as input and outputting the termination probabilities for every option. We keep a separate neural network for every low-level policy, each of which takes $s$ as input and outputs the parameters of a categorical over actions (we assume a discrete action space, but our method can easily be modified for continuous action spaces). Note that, other than the Bayesian treatment of $\eta$ and the inclusion of $\alpha$ in $\theta$, our model is identical to DDO. We will detail differences in how learning is performed in this section, and later show how our model can be made nonparametric.

As with CompILE, na\"ively doing behavioural cloning is intractable, and we therefore introduce an approximate posterior $q_\phi(\eta, \zeta|\xi)$. As we will cover in detail later, we endow this posterior with a structure respecting the conditional independences implied by Figure \ref{fig:graphical_model}, and unlike CompILE, we do not have to arbitrarily specify a number of segments.

\subsection{Objective}

We jointly train $\theta$ and $\phi$ by using the ELBO as the maximization objective, which is now given by:
\begin{align}\label{eq:obj}
    L(\theta, \phi) := & \; \mathbb{E}_{q_\phi(\eta, \zeta|\xi)}[\log p_\theta(\zeta, \xi|\eta) - \log q_\phi(\zeta|\eta, \xi)] \nonumber\\
    & - \mathbb{KL}(q_\phi(\eta)||p(\eta)) \leq \log p_\theta(\xi).
\end{align}
We use the GS distribution in order to perform the reparameterization trick and backpropagate through the ELBO. If we write the first term in more detail, we get:
\begin{align}
    \log p_\theta(\zeta,\xi|\eta) = \displaystyle \sum_{i=1}^N \log p_\theta(\zeta^{(i)}, \xi^{(i)}|\eta),
\end{align}
where each term is given by:
\begin{align}\label{joint}
    \medmath{\log p_\theta(\zeta^{(i)}, \xi^{(i)}|\eta)} & = \medmath{\log \rho(s_0^{(i)}) + \log \delta_{b_0^{(i)}=1}} \nonumber\\
    & \hspace{-1.5cm} + \displaystyle \medmath{\log \eta(h_0^{(i)}) + \sum_{t=1}^{T-1} \log p_\theta(b_t^{(i)}, h_t^{(i)}|h_{t-1}^{(i)}, s_t^{(i)}, \eta)} \\
    & \hspace{-1.5cm} + \medmath{\sum_{t=0}^{T-1} \log \pi_{h_t^{(i)}}(a_t^{(i)}|s_t^{(i)}) + \log P(s_{t+1}^{(i)}|s_t^{(i)}, a_t^{(i)})}.\nonumber
\end{align}
Note that the initial state distribution and environment dynamics terms cannot be evaluated but, as they do not depend on the parameters being optimized, they can be ignored. We can further decompose the term for $b_t^{(i)}$ and $h_t^{(i)}$:
\begin{align}
    p_\theta(b_t^{(i)}=1,h_t^{(i)}|h_{t-1}^{(i)}, s_t^{(i)}, \eta) = & \; \psi_{h_{t-1}^{(i)}}(s_t^{(i)})\eta(h_t^{(i)}) \label{b1}\\
    p_\theta(b_t^{(i)}=0,h_t^{(i)}|h_{t-1}^{(i)}, s_t^{(i)}, \eta) = & \; (1-\psi_{h_{t-1}^{(i)}}(s_t^{(i)})) \nonumber \\
    & \cdot \delta_{h_t^{(i)} = h_{t-1}^{(i)}}, \label{b0}
\end{align}
where $\{\psi_h\}_h$ are the termination functions. Note that equations \ref{joint}, \ref{b1} and \ref{b0} assume that the $b$'s are binary and the $h$'s are categorical (one-hot) through the Kronecker delta terms and evaluating $\eta$ at different $h$ terms. Since we use a Gumbel-Softmax relaxation to optimize the ELBO, the sampled values of $b$'s and $h$'s will be $(0,1)$ and simplex-valued, respectively. We therefore relax the corresponding terms so that they can be evaluated for such values and gradients can be propagated. We detail the relaxations in Appendix \ref{app:relaxation}.

\subsection{Variational Posterior}

We now describe the structure of our approximate posterior $q_\phi(\eta, \zeta|\xi)$. First, we observe from Figure \ref{fig:graphical_model} through the rules of $d$-separation \citep{koller2009probabilistic} that $\zeta^{(i)}$ is independent of $\zeta^{(j)}$ given $\eta$ and $\xi$, provided $i \neq j$. Additionally, given $\eta$, $\zeta^{(i)}$ depends on $\xi$ only through $\xi^{(i)}$. We thus take $q_\phi(\eta, \zeta|\xi)$ to obey the following conditional independence relationship:
\begin{align}
    q_\phi(\eta, \zeta|\xi) = q_\phi(\eta) \displaystyle \prod_{i=1}^N q_\phi(\zeta^{(i)}|\eta, \xi^{(i)}), \label{eqn:posterior1}
\end{align}
which holds for the true posterior as well.

Since $\eta$ is a global variable (i.e. it does not have components for each trajectory $i$) while $\zeta=\{\zeta^{(i)}\}_i$ is composed of local variables assigned to the corresponding $\xi^{(i)}$, we treat $\eta$ in a non-amortized \citep{gershman2014amortized} way (i.e. the parameters of $q_\phi(\eta)$ are directly optimized, there is no neural network), while we treat $\zeta^{(i)}$ in an amortized way (i.e. a neural network takes $\eta$ and $\xi^{(i)}$ as inputs to produce the parameters of a distribution).

As previously hinted, we parameterize $q_\phi(\eta)$ as a sequence of Kumaraswamy distributions to which the stick-breaking procedure is applied, which allows us to straightforwardly use the reparameterization trick.

As mentioned above, $q_\phi(\zeta^{(i)}|\eta, \xi^{(i)})$ is treated in an amortized way, so that a neural network takes $\eta$ and $\xi^{(i)}$ as input and produces the parameters for the distribution of $\zeta^{(i)}$, i.e. the Bernoulli parameters of each $b_t^{(i)}$ and categorical parameters of $h_t^{(i)}$ for $t=0,\dots,T-1$. We use an autoregressive structure for $q_\phi(\zeta^{(i)}|\eta, \xi^{(i)})$, and assume conditional independence of options and terminations at every time step:
\begin{align}\label{eq:autoreg_posterior}
    \medmath{q_\phi(\zeta^{(i)}|\eta, \xi^{(i)})} = & \; \medmath{q_\phi(b_0^{(i)} |\eta, \xi^{(i)}) q_\phi(h_0^{(i)} |\eta, \xi^{(i)})} \nonumber\\
    & \hspace{-2cm} \medmath{\cdot \prod_{t=1}^{T-1} q_\phi(b_t^{(i)} | \zeta_{t-1}^{(i)}, \eta, \xi^{(i)}_{t:T-1}) q_\phi(h_t^{(i)} | \zeta_{t-1}^{(i)}, \eta, \xi^{(i)}_{t:T-1})},
\end{align}
where $\xi_{t}^{(i)} = (s_{t}^{(i)}, a_{t}^{(i)})$ and $\zeta_{t}^{(i)} = (b_{t}^{(i)}, h_{t}^{(i)})$. Note that conditioned on $(\zeta_{t-1}^{(i)}, \eta, \xi^{(i)})$, $(b_t^{(i)}, h_t^{(i)})$ is independent of $\xi_{0:t-1}$, which can again be verified through Figure \ref{fig:graphical_model}. In other words, states and actions after time $t$ can be useful to infer the option and termination at time $t$; but previous states and actions are not (assuming the previous option and termination are known). This is why we only condition on $\xi^{(i)}_{t:T-1}$ in Equation \ref{eq:autoreg_posterior}, rather than on $\xi^{(i)}$.
In practice, we use an LSTM \citep{hochreiter1997long} to parse $\xi^{(i)}$ in reverse order, then sequentially sample $\zeta^{(i)}$ using multi-layer perceptrons (MLPs) to output the distributions' parameters as shown below (the MLPs' inputs are concatenated):
\begin{align}
        \medmath{b_{t}^{(i)} | \zeta_{t-1}^{(i)}, \eta, \xi^{(i)}} & \medmath{ \sim \text{\scriptsize{GS}}_\tau(\cdot | \text{\scriptsize{MLP}}_b(\text{\scriptsize{LSTM}}(\xi^{(i)}_{t:T-1}), \eta, \zeta_{t-1}^{(i)}))}, \label{encoder1}\\
        \medmath{h_{t}^{(i)} | \zeta_{t-1}^{(i)}, \eta, \xi^{(i)}} & \medmath{ \sim \text{\scriptsize{GS}}_\tau(\cdot | \text{\scriptsize{MLP}}_h(\text{\scriptsize{LSTM}}(\xi^{(i)}_{t:T-1}), \eta, \zeta_{t-1}^{(i)}))}. \label{encoder2}
\end{align}
The $b$ and $h$ indices reflect the fact that we use two separate output heads. The MLPs share their parameters except those of their last linear layer. The $\text{\small{LSTM}}(\xi^{(i)}_{t:T-1})$ term denotes the hidden state of the LSTM layer taken at time step $t$. Also note that we abuse notation and use $b_t^{(i)}$, $h_t^{(i)}$ and $\zeta_t^{(i)}$ for both the discrete variables and their relaxed counterparts. We refer to the LSTM and the MLP heads as the \textit{encoder}.

\subsection{Entropy Regularizer}\label{sec:entropy_reg}

In some preliminary experiments, we observed that our model could get stuck in some local optima where too few options were used. To address this issue, we add a regularizing term to the ELBO in Equation \ref{eq:obj}, defined as the entropy over the average of the sampled options $\{h_t^{(i)}\}$. More formally, we define $h^{(i)}_{avg} := \frac{1}{T}\sum_{t=0}^{T-1} h_t^{(i)}$, which is a probability vector of size $K$ and consider the following regularizing term that we want to maximize:
\begin{equation}\label{eq:entropy}
    l_{ent} :=  - \mathbb{E}_{q_\phi(\zeta|\xi)}\left[\sum_{i=1}^N \sum_{k=0}^{K-1} h^{(i)}_{avg,k}\log \left( h^{(i)}_{avg,k} \right)\right] ,
\end{equation}
where $h^{(i)}_{avg,k}$ is the $k$-th coordinate of $h^{(i)}_{avg}$. We also use the reparameterization trick to backpropagate through Equation \ref{eq:entropy}. This term encourages the model to use all the available options in equal amounts. Our final maximization objective is given by the following expression:
\begin{equation}
    L(\theta, \phi) + \lambda_{ent} l_{ent},
\end{equation}
where we anneal $\lambda_{ent} \geq 0$ throughout training with a fixed decay rate. We highlight that we only anneal $\lambda_{ent}$ since the ELBO is a highly principled objective, and annealing causes the objective being optimized to be closer and closer to the ELBO. That being said, we did not observe significant empirical differences between annealing and fixing $\lambda_{ent}$ throughout training (see Appendix \ref{app:extra_exp}).
\section{Nonparametric Models}

We now describe how to use a nonparametric prior to remove the need for a prespecified number of options.

\subsection{GEM Prior}\label{sec:gem_prior}

We now put a $\operatorname{GEM}(\alpha)$ prior over $\eta$, replacing the prior obtained from truncating the stick-breaking process at $K$ steps. The result is a nonparametric model which assumes a countably infinite number of options. Doing posterior inference over $\eta$ then enables learning the number of options present in the observed trajectories.

Note that $\theta$ should have an infinite number of parameters, since we consider an infinite number of options, but we detail below how we manage this in practice.

\subsection{Truncation} \label{sec:heuristic}

Recall that $\eta$ is now an infinite dimensional vector whose entries add up to one. A na\"ive attempt to optimize the objective from Equation \ref{eq:obj} would thus involve not only sampling infinitely long $\eta$'s, but also neural networks with infinitely many parameters (for terminations), or infinitely many neural networks (for low-level policies); this is clearly infeasible.
\citet{nalisnick2016stick} truncate $\eta$ and consider only the $K$ first coordinates, treating $K$ as a hyperparameter. We highlight that fixing $K$ in the approximate posterior yields the same objective as reverting the nonparametric $\operatorname{GEM}$ prior back to a $K$-dimensional prior; thus discarding the nonparametric aspect of the model. Even in this setting, our fixed-$K$ model remains a novel way of learning options offline. Nonetheless, in order to properly retain the nonparametric aspect of the model, we allow $K$ to increase throughout training. We check at regular intervals (every $n_K$ training steps) whether we should increment $K$ by looking at the ``usage'', $U(h)$, for each option $h$:
\begin{align}
    \medmath{U(h) := \frac{1}{NT}\sum_{i=1}^N \sum_{t=0}^{T-1} \mathds{1} \left(h = \underset{h_k:0 \leq k < K}{\operatorname{argmax}} \; \pi_{h_k}(a_t^{(i)} | s_t^{(i)}) \right)},
\end{align}
 i.e. $U(h)$ is the fraction of steps for which we inferred $h$ to be the option most likely to have been in use.
 Our rule for adding skills (clusters) is inspired by the small variance asymptotics Bayesian nonparametrics literature \citep{kulis2011revisiting, broderick2013mad}.
 We increase $K$ by $1$ if there is no option $h_k$ for $k \in \{0, ..., K-1\}$ such that $U(h_k) < \delta / K$ for a fixed hyperparameter $\delta < 1$.
  When increasing $K$, we add a new low-level policy, and expand the termination function by adding a new row to the last linear layer, so that the output changes from size $K$ to size $K+1$. We handle this new option in the encoder's MLPs (see Equations \ref{encoder1} and \ref{encoder2}) by adding two new columns to their first layer (to take into account the new input size, as both $\eta$ and $\zeta_{t-1}^{(i)}$ have increased in length by 1) and a new row to the last layer of $\text{MLP}_h$ (to output the correct number of parameters for $\zeta_t^{(i)}$). We also add parameters for another Kumaraswamy distribution to $\phi$. All new parameters are initialized randomly.
The logic behind this heuristic is that there is no need to add a new option (i.e. increment $K$) if there is an existing option which is rarely used.

\subsection{Nonparametric CompILE}\label{sec:npcomp}

Finally, we show that CompILE can now be easily turned into a nonparametric model thanks to the machinery we have developed. When placing a $\operatorname{GEM}(\alpha)$ prior over $\eta$, the CompILE ELBO from Equation \ref{eq:comp} changes to accommodate the fact that $\eta$ is now being treated in a Bayesian way rather than being fixed and uniform. The ELBO then becomes:
\begin{align}\label{eq:obj2}
    L(\theta, \phi) := & \; \mathbb{E}_{q_\phi(\eta, \zeta|\xi)}[\log p_\theta(\zeta, \xi|\eta) - \log q_\phi(\zeta|\eta, \xi)] \nonumber\\
    & - \mathbb{KL}(q_\phi(\eta)||p(\eta)).
\end{align}
While this objective looks identical to ours, we highlight once again that we have switched back to CompILE notation, and that the terms are not identical to those of Equation \ref{eq:obj}. The only practical changes to CompILE are then: $(1)$ that we also use a non-amortized posterior $q_{\phi}(\eta)$ obtained by stick-breaking Kumaraswamy random variables, and so the $\mathbb{KL}$ term in the ELBO is identical to that of Equation \ref{eq:obj}; $(2)$ the approximate posterior becomes $q_\phi(\eta, \zeta|\xi) = q_\phi(\eta)q_\phi(\zeta|\eta, \xi)$, which implies that, as with our own model, the neural network modeling $q_\phi(\zeta|\eta, \xi)$ now needs to take $\eta$ as an extra input, although we leave the rest of the structure used in CompILE intact; and $(3)$ when evaluating $\eta(h_j^{(i)})$ for $p_\theta(\zeta, \xi|\eta)$,  where $h_j^{(i)}$ is the skill used for the $j$-th segment in the $i$-th trajectory, we actually evaluate $\eta$ (as we do in options), rather than using $1/K$ as implied by the uniform distribution. Additionally, for the nonparametric version of CompILE we use the same truncation and entropy regularizer as we used for our options model.

\section{Experiments}\label{sec:exps}

\begin{figure}
    \centering
    \includegraphics[width=\linewidth]{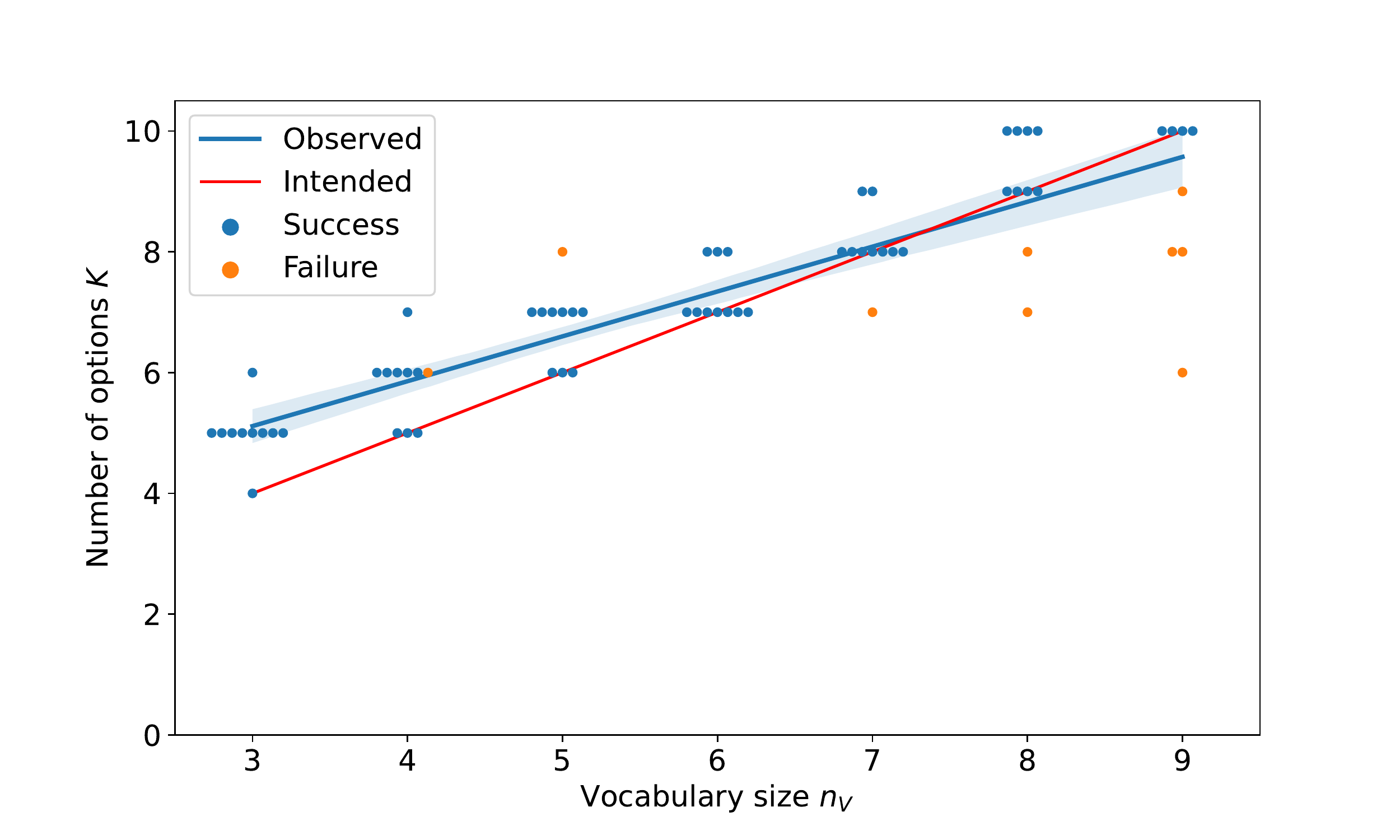}
    \caption{Results for our proof-of-concept environment. We consider 10 random seeds per vocabulary size $n_V$, and a run is considered a success if one of the learned options selects the correct action at $t=4$ with probability at least $0.95$. The red line shows the number of options that should be recovered ($n_V +1$), while the blue line is the linear regression of the actual recovered numbers and closely mimics the red line.}
    \label{fig:toy_env_results}
\end{figure}

\begin{figure*}[th!]
    \centering
    \begin{subfigure}[b]{0.45\textwidth}
        \centering
         \includegraphics[width=\textwidth]{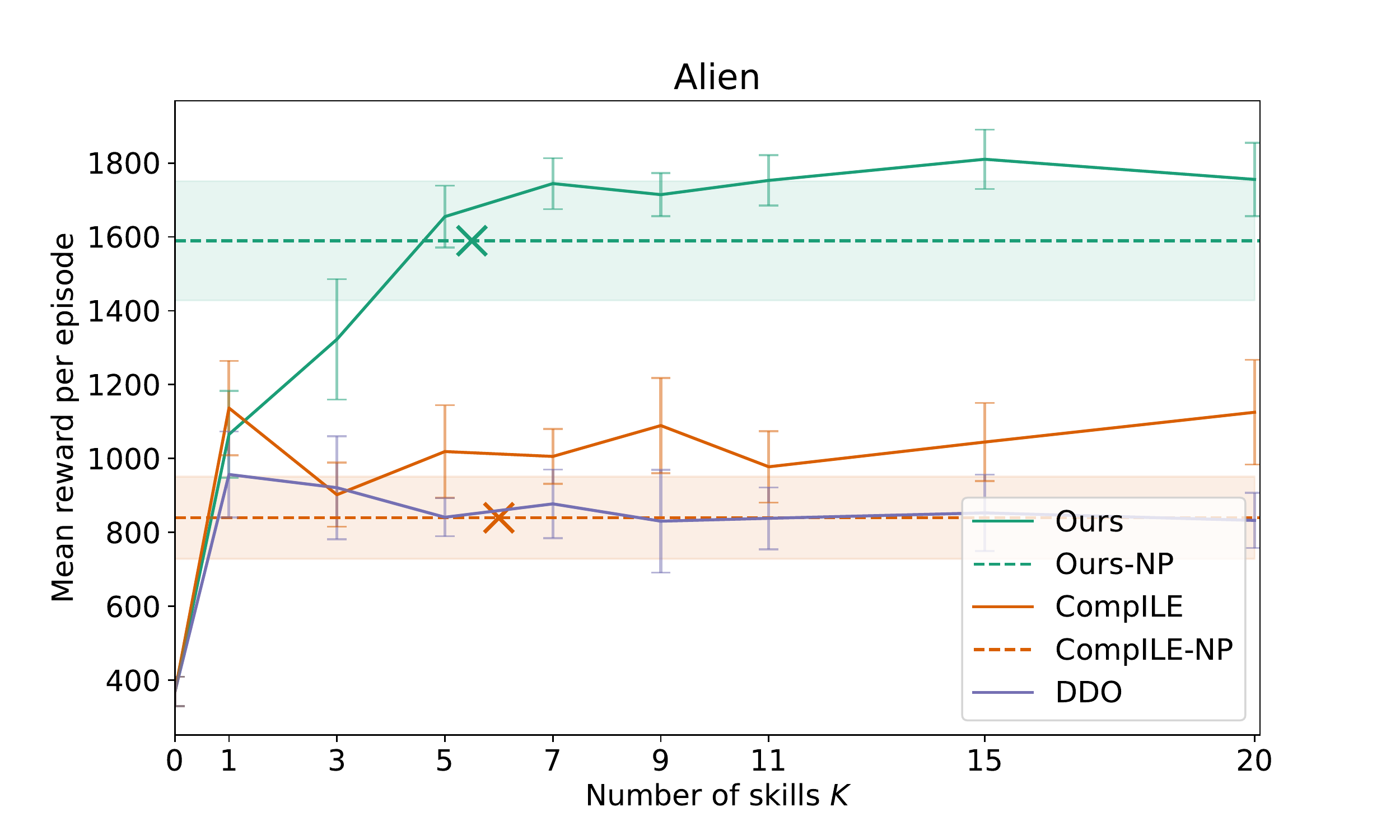}
    \end{subfigure}
    \begin{subfigure}[b]{0.45\textwidth}
        \centering
         \includegraphics[width=\textwidth]{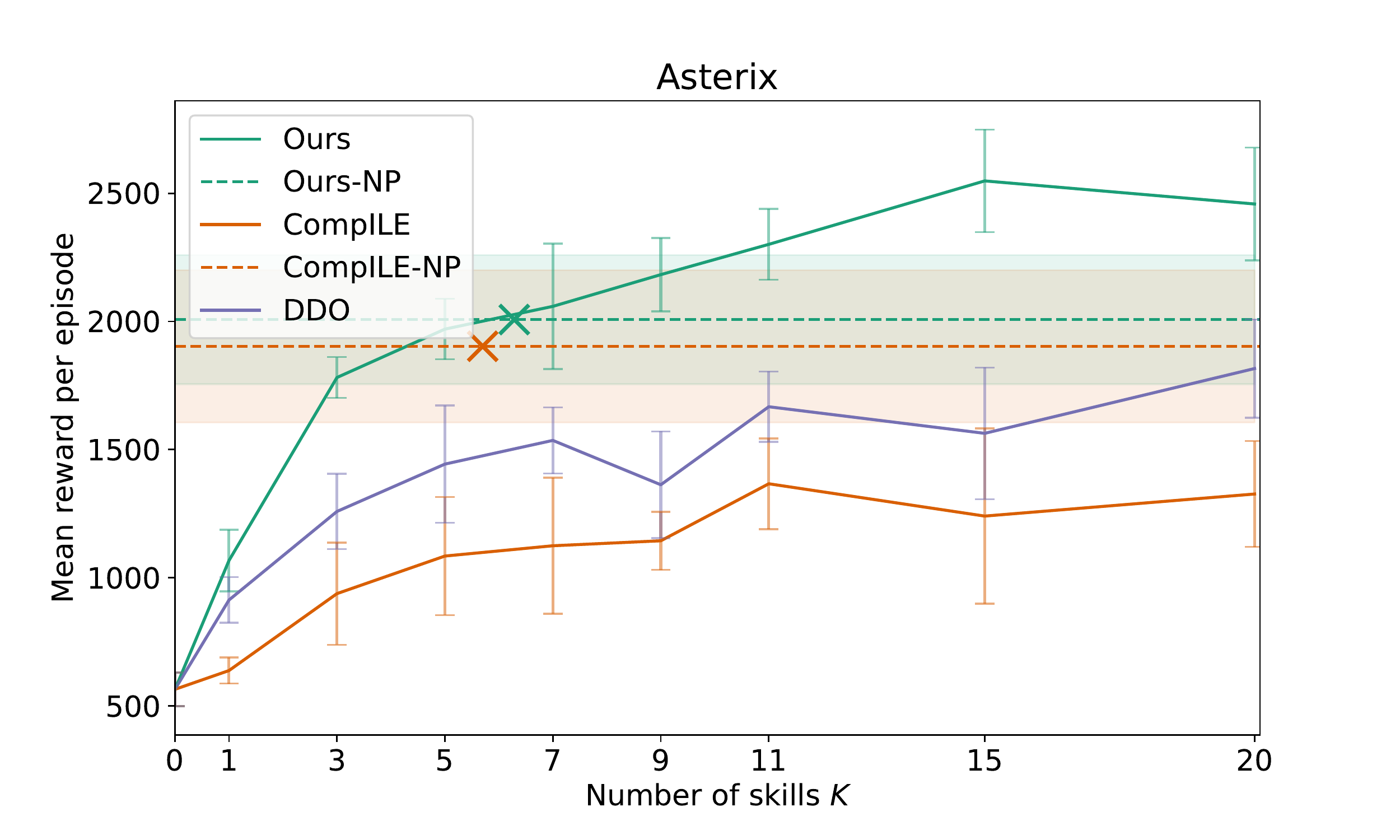}
    \end{subfigure}
    \begin{subfigure}[b]{0.45\textwidth}
        \centering
         \includegraphics[width=\textwidth]{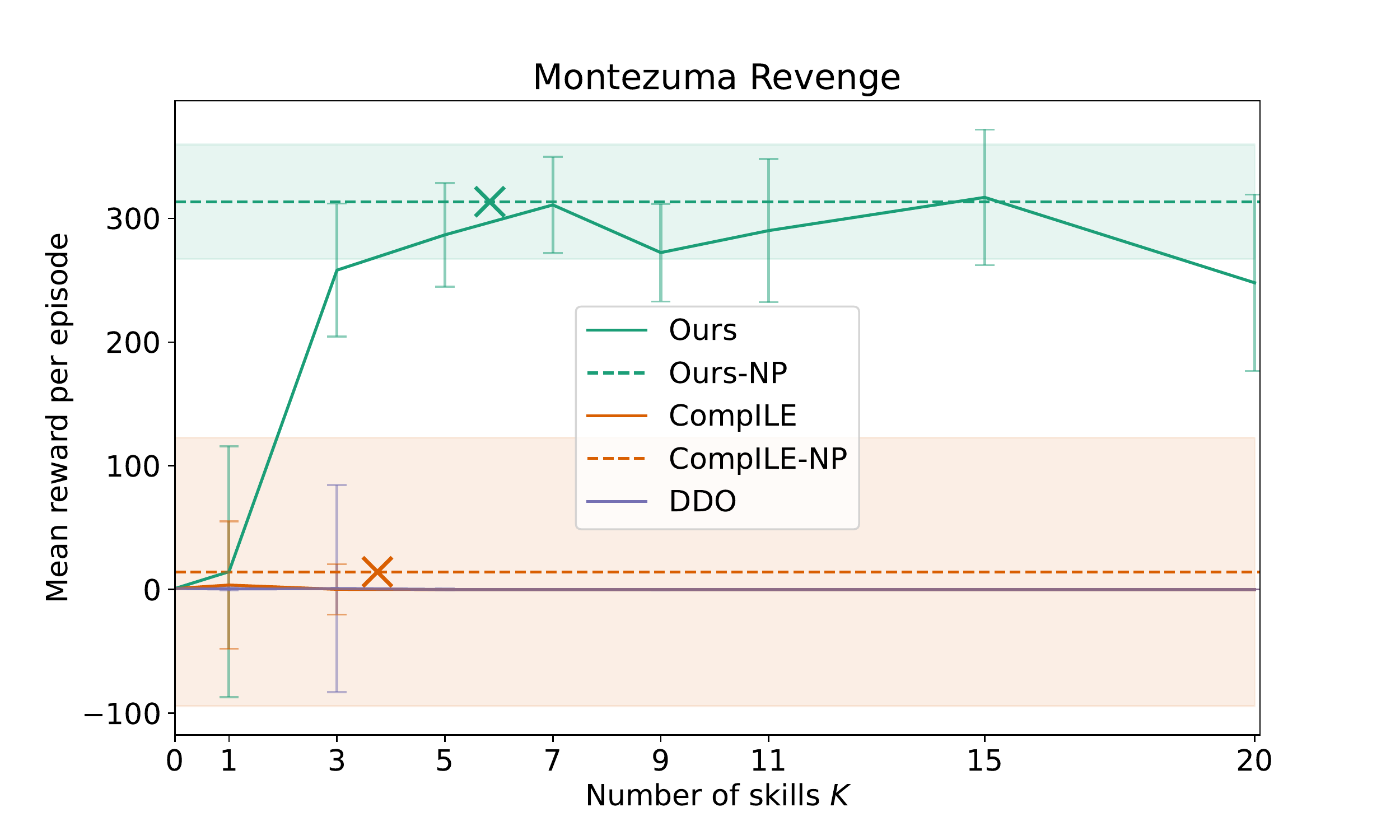}
    \end{subfigure}
    \begin{subfigure}[b]{0.45\textwidth}
        \centering
         \includegraphics[width=\textwidth]{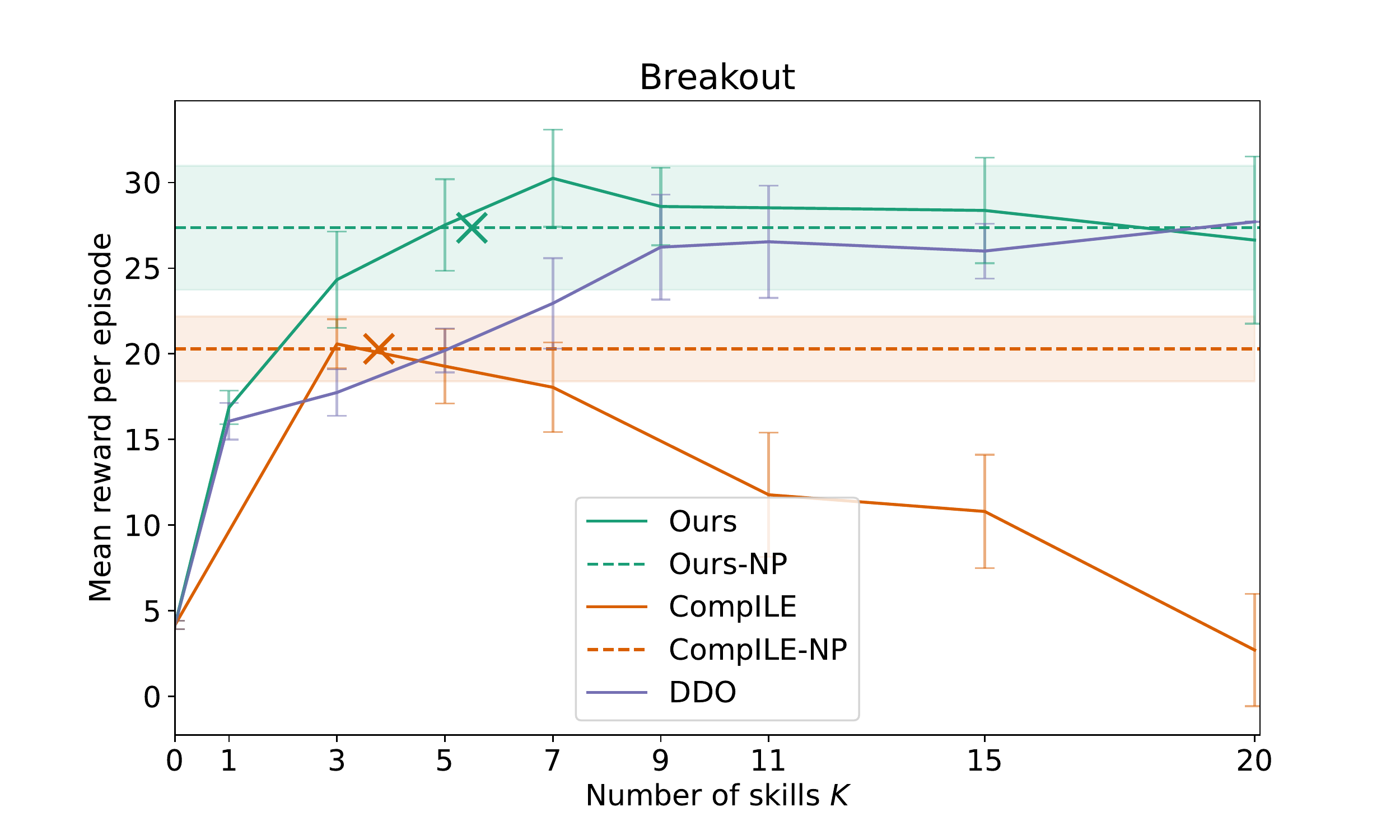}
    \end{subfigure}
    \begin{subfigure}[b]{0.45\textwidth}
        \centering
         \includegraphics[width=\textwidth]{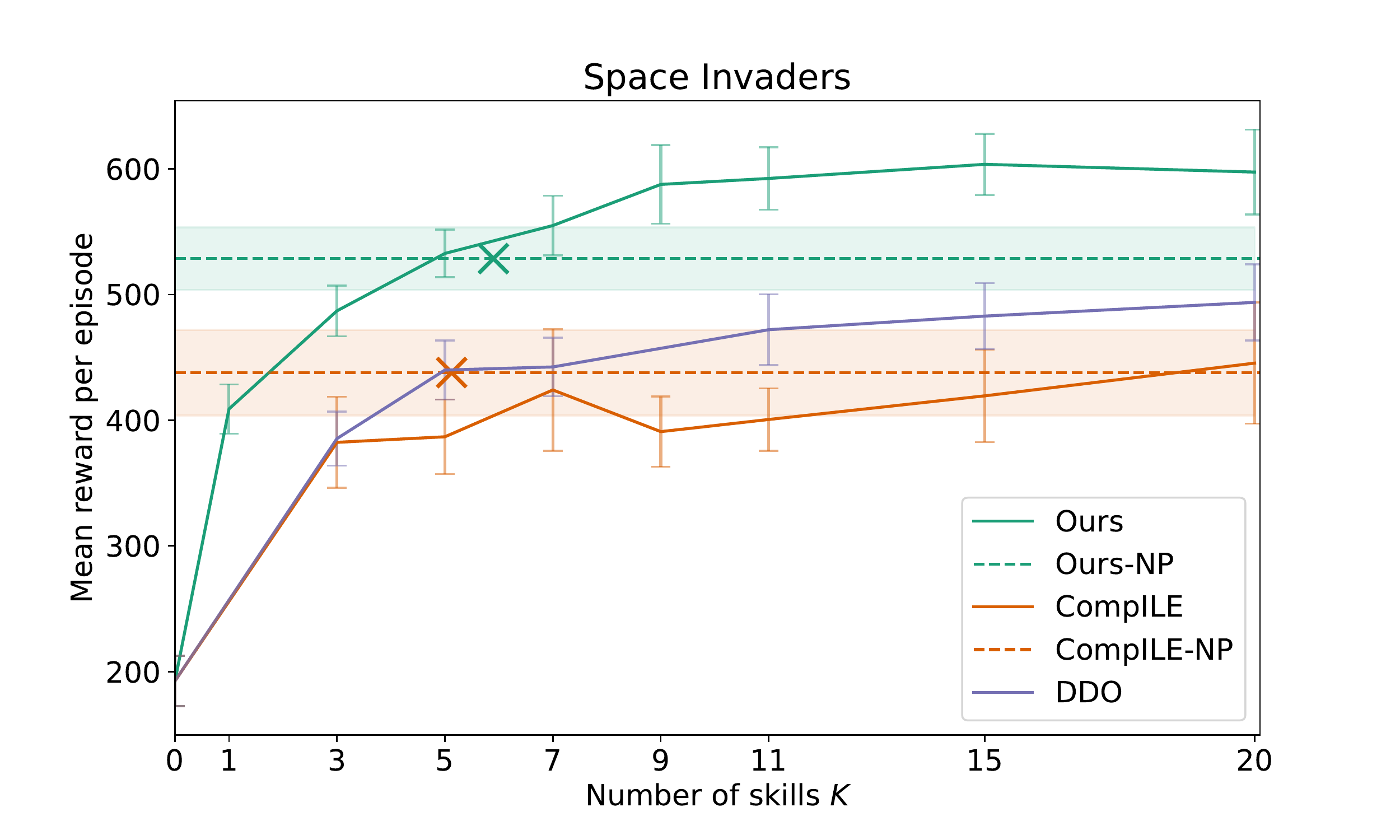}
    \end{subfigure}
    \begin{subfigure}[b]{0.45\textwidth}
        \centering
         \includegraphics[width=\textwidth]{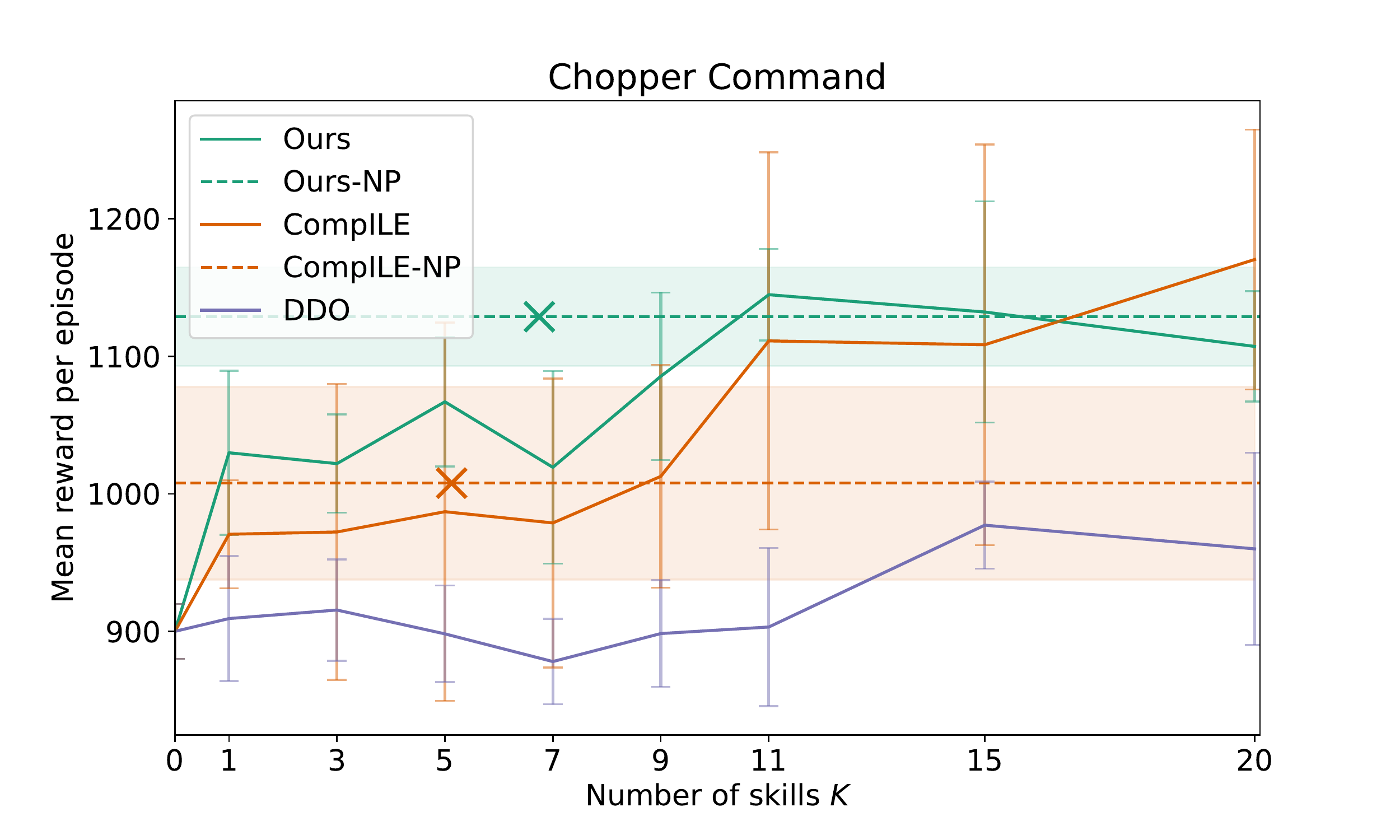}
    \end{subfigure}
    \caption{Results for Atari environments showing mean reward per episode (after training) averaged across at least $10$ seeds for different numbers of skills, with corresponding standard errors. Dashed lines correspond to nonparametric models where $K$ is not specified. The label `ours' denotes the fixed-$K$ version of our options model, and the `-NP' suffix denotes nonparametric versions, both for ours and CompILE.}
    \label{fig:atari_exp_results}
\end{figure*}

The goal of our experiments is twofold: to show that our options framework learns more useful skills than DDO and CompILE, and also that the nonparametric extensions of our own model and CompILE (which circumvent the need to specify $K$) match the performance of their respective parametric versions with $K$ tuned as a hyperparameter. The former goal highlights the usefulness of incorporating variational inference advances to offline option learning, and the latter highlights the benefits of using Bayesian nonparametrics for skill discovery.  All experimental details are given in Appendix \ref{app:exp_details}.

\subsection{Proof-of-Concept Environment}

We evaluate our model in an environment we designed, where we know the number of skills required to fully reconstruct the expert trajectories. Our nonparametric model recovers sufficient options for reconstruction, without this number being specified beforehand. In this environment, an agent receives a message $m$ from a vocabulary $\{0, ..., n_{V}-1\}$ at time $t=0$ and has to emit the same message $m$ at time $t=4$. The agent's observation is $(t, m)$ at $t=0$ and $(t, -1)$ for all subsequent timesteps, and they can take a discrete action $a_t \in \{0, ..., n_V-1\}$ at each timestep $t$. The task is considered a success if $a_{4} = m$. Note that this task cannot be solved by a Markovian policy, since at time $t=4$ the agent needs to remember the message they received at time $t=0$. However, it is possible to solve the environment with a hierarchical policy with at least $n_V$ options: if every possible message has a corresponding option whose low-level policy emits that message at time $t=4$, then an optimal agent needs just select the appropriate option at time $t=0$ when observing $m$ and not terminate before time $t=4$. We generate expert trajectories for various vocabulary sizes. Considering the heuristic we use to update $K$ (see Section \ref{sec:heuristic}), we expect our method to recover at least $K = n_V + 1$ ($K$ options equally used, and one unused option). We show the results in Figure \ref{fig:toy_env_results}: indeed, we recover enough options without overestimating $K$ by more than $1$. We also perform ablations in Appendix \ref{app:extra_exp}, where we show that removing the entropy regularizer from Section \ref{sec:entropy_reg} significantly hurts performance.

\subsection{Atari Environments}

\begin{figure}[t]
    \centering
    \includegraphics[width=0.45\textwidth]{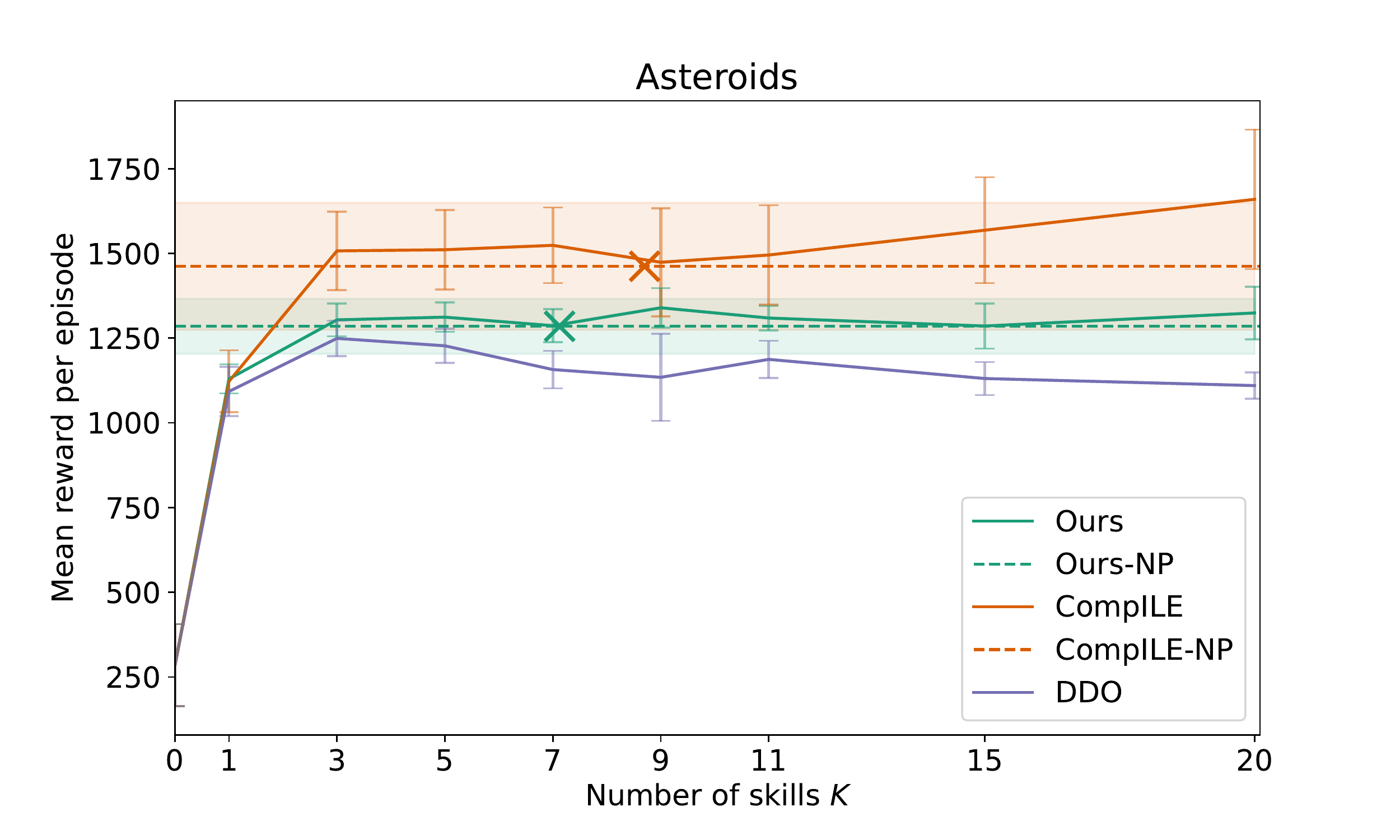}
    \caption{Additional Atari environment, same setup as in Figure \ref{fig:atari_exp_results}.}
    \label{fig:atari_exp_results2}
\end{figure}

We further test our model on several games from the Atari learning environment \citep{bellemare2013arcade}.
For each game, we use expert trajectories generated by a trained Ape-X agent \citep{horgan2018distributed, such2019atari}. To evaluate the quality of the discovered options, we train a PPO agent \citep{schulman2017proximal} on an environment for which the action space has been `augmented' using the skills, similarly to \citet{fox2017multi}. In this augmented environment, alongside the default game actions, the agent can choose to perform `enhanced actions', which consist of acting according to one of the learned skills until it terminates. We modify the PPO algorithm to take these enhanced actions into account (see Appendix \ref{app:augmented}). We compare our method to DDO and CompILE. We compare both the parametric version of our model where $K$ is treated as a hyperparameter, and the nonparametric versions of our model and CompILE. Results are summarized in Figures \ref{fig:atari_exp_results} and \ref{fig:atari_exp_results2}. We can see that $(1)$ our model learns more useful skills than DDO and CompILE (green lines consistently above purple and orange ones, except for the Asteroids environment), and that $(2)$ the nonparametric version of our model matches the performance of tuning $K$ (dashed green lines match the peak of the green lines for all environments, except Asterix and Space Invaders), and similarly for CompILE's nonparametric version (analogously for orange lines, except for Alien and Chopper Command this time). Note that the performance of the agents in the original (i.e. not augmented) versions of the environments correspond to the `$K=0$' entries in the figures: the significant improvements with $K>0$ highlight the relevance of offline skill discovery.  The crosses in Figures \ref{fig:atari_exp_results} and \ref{fig:atari_exp_results2} show the recovered values of $K$ (averaged across runs) for the nonparametric models, and we can see that, except for a few cases, the number of recovered options closely matches the smallest optimal number that would be obtained by tuning $K$ through many runs, underlining the benefits of our nonparametric approaches. We include visualizations of our learned options in the github repository containing our code.\footnote{\url{https://github.com/layer6ai-labs/BNPO/blob/main/visualization/visualization.md}}

We highlight that we did not tune $\lambda_{ent}$ nor its decay schedule and used the same settings across all experiments, so that not tuning $K$ does not come at the cost of having to tune other parameters. We present ablations with respect to the entropy regularizer in Appendix \ref{app:extra_exp}, where we found that while it did not hurt performance, this regularizer was not as fundamental as in the proof-of-concept environment. We also include in Appendix \ref{app:extra_exp} ablations over trajectory length, and find that while results vary for all methods, ours remains the strongest regardless of trajectory length.

\section{Conclusions and Future Work}

We introduced a novel approach for offline option discovery taking advantage of variational inference and the Gumbel-Softmax distribution. We leave an exploration of continuous relaxation \citep{kool2020estimating,potapczynski2020invertible,paulus2020rao} and variational inference advances \citep{burda2015importance, roeder2017sticking} within our framework for future research.

We also highlighted an unexplored connection between skill discovery and Bayesian nonparametrics, and showed that specifying $K$ as a hyperparameter can be avoided by placing a $\operatorname{GEM}$ prior over skills, both in our model and other skill discovery frameworks. We hope our work will motivate further research combining reinforcement learning and Bayesian nonparametrics, for example through small-variance asymptotics \citep{kulis2011revisiting,jiang2012small,broderick2013mad,roychowdhury2013small} for hard rather than soft clustering of skills, or with hierarchical models \citep{teh2006hierarchical}. Finally, we also hope that the technique we developed to increase $K$ throughout training will find uses in other Bayesian nonparametric models, outside of skill discovery.


\section*{Acknowledgements}
We thank the anonymous reviewers for their feedback, which helped improve our paper. We also thank Junfeng Wen for a useful suggestion on an ablation experiment. Our code was written in Python \citep{vanrossum1995python,oliphant2007python} and particularly relied on the following packages: Matplotlib \citep{Hunter:2007}, TensorFlow \citep{tensorflow2015-whitepaper} (in particular for TensorBoard), gym \citep{brockman2016openai}, Jupyter Notebook \citep{Kluyver2016jupyter}, PyTorch \citep{paszke2019pytorch}, NumPy \citep{harris2020array}, and Stable-Baselines3 \citep{stable-baselines3}.

\bibliography{main_bib}

\begin{thebibliography}{69}
\providecommand{\natexlab}[1]{#1}
\providecommand{\url}[1]{\texttt{#1}}
\expandafter\ifx\csname urlstyle\endcsname\relax
  \providecommand{\doi}[1]{doi: #1}\else
  \providecommand{\doi}{doi: \begingroup \urlstyle{rm}\Url}\fi

\bibitem[Abadi et~al.(2015)Abadi, Agarwal, Barham, Brevdo, Chen, Citro,
  Corrado, Davis, Dean, Devin, Ghemawat, Goodfellow, Harp, Irving, Isard, Jia,
  Jozefowicz, Kaiser, Kudlur, Levenberg, Man\'{e}, Monga, Moore, Murray, Olah,
  Schuster, Shlens, Steiner, Sutskever, Talwar, Tucker, Vanhoucke, Vasudevan,
  Vi\'{e}gas, Vinyals, Warden, Wattenberg, Wicke, Yu, and
  Zheng]{tensorflow2015-whitepaper}
Abadi, M., Agarwal, A., Barham, P., Brevdo, E., Chen, Z., Citro, C., Corrado,
  G.~S., Davis, A., Dean, J., Devin, M., Ghemawat, S., Goodfellow, I., Harp,
  A., Irving, G., Isard, M., Jia, Y., Jozefowicz, R., Kaiser, L., Kudlur, M.,
  Levenberg, J., Man\'{e}, D., Monga, R., Moore, S., Murray, D., Olah, C.,
  Schuster, M., Shlens, J., Steiner, B., Sutskever, I., Talwar, K., Tucker, P.,
  Vanhoucke, V., Vasudevan, V., Vi\'{e}gas, F., Vinyals, O., Warden, P.,
  Wattenberg, M., Wicke, M., Yu, Y., and Zheng, X.
\newblock {TensorFlow}: Large-scale machine learning on heterogeneous systems,
  2015.
\newblock URL \url{https://www.tensorflow.org/}.
\newblock Software available from tensorflow.org.

\bibitem[Achiam et~al.(2018)Achiam, Edwards, Amodei, and
  Abbeel]{achiam2018variational}
Achiam, J., Edwards, H., Amodei, D., and Abbeel, P.
\newblock Variational option discovery algorithms.
\newblock \emph{arXiv preprint arXiv:1807.10299}, 2018.

\bibitem[Ajay et~al.(2021)Ajay, Kumar, Agrawal, Levine, and
  Nachum]{ajay2021opal}
Ajay, A., Kumar, A., Agrawal, P., Levine, S., and Nachum, O.
\newblock {\{}OPAL{\}}: Offline primitive discovery for accelerating offline
  reinforcement learning.
\newblock In \emph{International Conference on Learning Representations}, 2021.
\newblock URL \url{https://openreview.net/forum?id=V69LGwJ0lIN}.

\bibitem[Argall et~al.(2009)Argall, Chernova, Veloso, and
  Browning]{argall2009survey}
Argall, B.~D., Chernova, S., Veloso, M., and Browning, B.
\newblock A survey of robot learning from demonstration.
\newblock \emph{Robotics and autonomous systems}, 57\penalty0 (5):\penalty0
  469--483, 2009.

\bibitem[Bacon et~al.(2017)Bacon, Harb, and Precup]{bacon2017option}
Bacon, P.-L., Harb, J., and Precup, D.
\newblock The option-critic architecture.
\newblock In \emph{Proceedings of the AAAI Conference on Artificial
  Intelligence}, volume~31, 2017.

\bibitem[Bellemare et~al.(2013)Bellemare, Naddaf, Veness, and
  Bowling]{bellemare2013arcade}
Bellemare, M.~G., Naddaf, Y., Veness, J., and Bowling, M.
\newblock The arcade learning environment: An evaluation platform for general
  agents.
\newblock \emph{Journal of Artificial Intelligence Research}, 47:\penalty0
  253--279, 2013.

\bibitem[Brockman et~al.(2016)Brockman, Cheung, Pettersson, Schneider,
  Schulman, Tang, and Zaremba]{brockman2016openai}
Brockman, G., Cheung, V., Pettersson, L., Schneider, J., Schulman, J., Tang,
  J., and Zaremba, W.
\newblock Openai gym.
\newblock \emph{arXiv preprint arXiv:1606.01540}, 2016.

\bibitem[Broderick et~al.(2013)Broderick, Kulis, and Jordan]{broderick2013mad}
Broderick, T., Kulis, B., and Jordan, M.
\newblock Mad-bayes: Map-based asymptotic derivations from bayes.
\newblock In \emph{International Conference on Machine Learning}, pp.\
  226--234. PMLR, 2013.

\bibitem[Burda et~al.(2015)Burda, Grosse, and
  Salakhutdinov]{burda2015importance}
Burda, Y., Grosse, R., and Salakhutdinov, R.
\newblock Importance weighted autoencoders.
\newblock \emph{arXiv preprint arXiv:1509.00519}, 2015.

\bibitem[Caron et~al.(2007)Caron, Davy, and Doucet]{caron2007generalized}
Caron, F., Davy, M., and Doucet, A.
\newblock Generalized polya urn for time-varying dirichlet process mixtures.
\newblock In \emph{Proceedings of the Twenty-Third Conference on Uncertainty in
  Artificial Intelligence}, pp.\  33--40, 2007.

\bibitem[Dempster et~al.(1977)Dempster, Laird, and Rubin]{dempster1977maximum}
Dempster, A.~P., Laird, N.~M., and Rubin, D.~B.
\newblock Maximum likelihood from incomplete data via the em algorithm.
\newblock \emph{Journal of the Royal Statistical Society: Series B
  (Methodological)}, 39\penalty0 (1):\penalty0 1--22, 1977.

\bibitem[Dietterich et~al.(1998)]{dietterich1998maxq}
Dietterich, T.~G. et~al.
\newblock The maxq method for hierarchical reinforcement learning.
\newblock In \emph{ICML}, volume~98, pp.\  118--126. Citeseer, 1998.

\bibitem[Dunson \& Xing(2009)Dunson and Xing]{dunson2009nonparametric}
Dunson, D.~B. and Xing, C.
\newblock Nonparametric bayes modeling of multivariate categorical data.
\newblock \emph{Journal of the American Statistical Association}, 104\penalty0
  (487):\penalty0 1042--1051, 2009.

\bibitem[Esmaili et~al.(1995)Esmaili, Sammut, and
  Shirazi]{esmaili1995behavioural}
Esmaili, N., Sammut, C., and Shirazi, G.
\newblock Behavioural cloning in control of a dynamic system.
\newblock In \emph{1995 IEEE International Conference on Systems, Man and
  Cybernetics. Intelligent Systems for the 21st Century}, volume~3, pp.\
  2904--2909. IEEE, 1995.

\bibitem[Eysenbach et~al.(2018)Eysenbach, Gupta, Ibarz, and
  Levine]{eysenbach2018diversity}
Eysenbach, B., Gupta, A., Ibarz, J., and Levine, S.
\newblock Diversity is all you need: Learning skills without a reward function.
\newblock \emph{arXiv preprint arXiv:1802.06070}, 2018.

\bibitem[Ferguson(1973)]{ferguson1973bayesian}
Ferguson, T.~S.
\newblock A bayesian analysis of some nonparametric problems.
\newblock \emph{The annals of statistics}, pp.\  209--230, 1973.

\bibitem[Florensa et~al.(2017)Florensa, Duan, and
  Abbeel]{florensa2017stochastic}
Florensa, C., Duan, Y., and Abbeel, P.
\newblock Stochastic neural networks for hierarchical reinforcement learning.
\newblock \emph{arXiv preprint arXiv:1704.03012}, 2017.

\bibitem[Fox et~al.(2017)Fox, Krishnan, Stoica, and Goldberg]{fox2017multi}
Fox, R., Krishnan, S., Stoica, I., and Goldberg, K.
\newblock Multi-level discovery of deep options.
\newblock \emph{arXiv preprint arXiv:1703.08294}, 2017.

\bibitem[Gershman \& Goodman(2014)Gershman and Goodman]{gershman2014amortized}
Gershman, S. and Goodman, N.
\newblock Amortized inference in probabilistic reasoning.
\newblock In \emph{Proceedings of the annual meeting of the cognitive science
  society}, volume~36, 2014.

\bibitem[Glynn(1990)]{glynn1990likelihood}
Glynn, P.~W.
\newblock Likelihood ratio gradient estimation for stochastic systems.
\newblock \emph{Communications of the ACM}, 33\penalty0 (10):\penalty0 75--84,
  1990.

\bibitem[Gregor et~al.(2016)Gregor, Rezende, and
  Wierstra]{gregor2016variational}
Gregor, K., Rezende, D.~J., and Wierstra, D.
\newblock Variational intrinsic control.
\newblock \emph{arXiv preprint arXiv:1611.07507}, 2016.

\bibitem[Harris et~al.(2020)Harris, Millman, van~der Walt, Gommers, Virtanen,
  Cournapeau, Wieser, Taylor, Berg, Smith, Kern, Picus, Hoyer, van Kerkwijk,
  Brett, Haldane, del R{\'{i}}o, Wiebe, Peterson, G{\'{e}}rard-Marchant,
  Sheppard, Reddy, Weckesser, Abbasi, Gohlke, and Oliphant]{harris2020array}
Harris, C.~R., Millman, K.~J., van~der Walt, S.~J., Gommers, R., Virtanen, P.,
  Cournapeau, D., Wieser, E., Taylor, J., Berg, S., Smith, N.~J., Kern, R.,
  Picus, M., Hoyer, S., van Kerkwijk, M.~H., Brett, M., Haldane, A., del
  R{\'{i}}o, J.~F., Wiebe, M., Peterson, P., G{\'{e}}rard-Marchant, P.,
  Sheppard, K., Reddy, T., Weckesser, W., Abbasi, H., Gohlke, C., and Oliphant,
  T.~E.
\newblock Array programming with {NumPy}.
\newblock \emph{Nature}, 585\penalty0 (7825):\penalty0 357--362, September
  2020.
\newblock \doi{10.1038/s41586-020-2649-2}.
\newblock URL \url{https://doi.org/10.1038/s41586-020-2649-2}.

\bibitem[Ho \& Ermon(2016)Ho and Ermon]{ho2016generative}
Ho, J. and Ermon, S.
\newblock Generative adversarial imitation learning.
\newblock \emph{Advances in neural information processing systems},
  29:\penalty0 4565--4573, 2016.

\bibitem[Hochreiter \& Schmidhuber(1997)Hochreiter and
  Schmidhuber]{hochreiter1997long}
Hochreiter, S. and Schmidhuber, J.
\newblock Long short-term memory.
\newblock \emph{Neural computation}, 9:\penalty0 1735--1780, 1997.

\bibitem[Horgan et~al.(2018)Horgan, Quan, Budden, Barth-Maron, Hessel,
  Van~Hasselt, and Silver]{horgan2018distributed}
Horgan, D., Quan, J., Budden, D., Barth-Maron, G., Hessel, M., Van~Hasselt, H.,
  and Silver, D.
\newblock Distributed prioritized experience replay.
\newblock \emph{arXiv preprint arXiv:1803.00933}, 2018.

\bibitem[Hunter(2007)]{Hunter:2007}
Hunter, J.~D.
\newblock Matplotlib: A 2d graphics environment.
\newblock \emph{Computing in Science \& Engineering}, 9\penalty0 (3):\penalty0
  90--95, 2007.
\newblock \doi{10.1109/MCSE.2007.55}.

\bibitem[Hussein et~al.(2017)Hussein, Gaber, Elyan, and
  Jayne]{hussein2017imitation}
Hussein, A., Gaber, M.~M., Elyan, E., and Jayne, C.
\newblock Imitation learning: A survey of learning methods.
\newblock \emph{ACM Computing Surveys (CSUR)}, 50\penalty0 (2):\penalty0 1--35,
  2017.

\bibitem[Ishwaran \& James(2001)Ishwaran and James]{ishwaran2001gibbs}
Ishwaran, H. and James, L.~F.
\newblock Gibbs sampling methods for stick-breaking priors.
\newblock \emph{Journal of the American Statistical Association}, 96\penalty0
  (453):\penalty0 161--173, 2001.

\bibitem[Jang et~al.(2016)Jang, Gu, and Poole]{jang2016categorical}
Jang, E., Gu, S., and Poole, B.
\newblock Categorical reparameterization with gumbel-softmax.
\newblock \emph{arXiv preprint arXiv:1611.01144}, 2016.

\bibitem[Jiang et~al.(2012)Jiang, Kulis, and Jordan]{jiang2012small}
Jiang, K., Kulis, B., and Jordan, M.
\newblock Small-variance asymptotics for exponential family dirichlet process
  mixture models.
\newblock \emph{Advances in Neural Information Processing Systems},
  25:\penalty0 3158--3166, 2012.

\bibitem[Khetarpal et~al.(2020)Khetarpal, Klissarov, Chevalier-Boisvert, Bacon,
  and Precup]{khetarpal2020options}
Khetarpal, K., Klissarov, M., Chevalier-Boisvert, M., Bacon, P.-L., and Precup,
  D.
\newblock Options of interest: Temporal abstraction with interest functions.
\newblock In \emph{Proceedings of the AAAI Conference on Artificial
  Intelligence}, volume~34, pp.\  4444--4451, 2020.

\bibitem[Kingma \& Ba(2014)Kingma and Ba]{kingma2014adam}
Kingma, D.~P. and Ba, J.
\newblock Adam: A method for stochastic optimization.
\newblock \emph{arXiv preprint arXiv:1412.6980}, 2014.

\bibitem[Kingma \& Welling(2013)Kingma and Welling]{kingma2013auto}
Kingma, D.~P. and Welling, M.
\newblock Auto-encoding variational bayes.
\newblock \emph{arXiv preprint arXiv:1312.6114}, 2013.

\bibitem[Kipf et~al.(2019)Kipf, Li, Dai, Zambaldi, Sanchez-Gonzalez,
  Grefenstette, Kohli, and Battaglia]{kipf2019compile}
Kipf, T., Li, Y., Dai, H., Zambaldi, V., Sanchez-Gonzalez, A., Grefenstette,
  E., Kohli, P., and Battaglia, P.
\newblock Compile: Compositional imitation learning and execution.
\newblock In \emph{International Conference on Machine Learning}, pp.\
  3418--3428. PMLR, 2019.

\bibitem[Kluyver et~al.(2016)Kluyver, Ragan-Kelley, P{\'e}rez, Granger,
  Bussonnier, Frederic, Kelley, Hamrick, Grout, Corlay, Ivanov, Avila, Abdalla,
  and Willing]{Kluyver2016jupyter}
Kluyver, T., Ragan-Kelley, B., P{\'e}rez, F., Granger, B., Bussonnier, M.,
  Frederic, J., Kelley, K., Hamrick, J., Grout, J., Corlay, S., Ivanov, P.,
  Avila, D., Abdalla, S., and Willing, C.
\newblock Jupyter notebooks -- a publishing format for reproducible
  computational workflows.
\newblock In Loizides, F. and Schmidt, B. (eds.), \emph{Positioning and Power
  in Academic Publishing: Players, Agents and Agendas}, pp.\  87 -- 90. IOS
  Press, 2016.

\bibitem[Koller \& Friedman(2009)Koller and Friedman]{koller2009probabilistic}
Koller, D. and Friedman, N.
\newblock \emph{Probabilistic graphical models: principles and techniques}.
\newblock MIT press, 2009.

\bibitem[Kool et~al.(2020)Kool, van Hoof, and Welling]{kool2020estimating}
Kool, W., van Hoof, H., and Welling, M.
\newblock Estimating gradients for discrete random variables by sampling
  without replacement.
\newblock \emph{arXiv preprint arXiv:2002.06043}, 2020.

\bibitem[Krishnan et~al.(2018)Krishnan, Garg, Patil, Lea, Hager, Abbeel, and
  Goldberg]{krishnan2018transition}
Krishnan, S., Garg, A., Patil, S., Lea, C., Hager, G., Abbeel, P., and
  Goldberg, K.
\newblock Transition state clustering: Unsupervised surgical trajectory
  segmentation for robot learning.
\newblock In \emph{Robotics Research}, pp.\  91--110. Springer, 2018.

\bibitem[Krishnan et~al.(2019)Krishnan, Garg, Liaw, Thananjeyan, Miller,
  Pokorny, and Goldberg]{krishnan2019swirl}
Krishnan, S., Garg, A., Liaw, R., Thananjeyan, B., Miller, L., Pokorny, F.~T.,
  and Goldberg, K.
\newblock Swirl: A sequential windowed inverse reinforcement learning algorithm
  for robot tasks with delayed rewards.
\newblock \emph{The International Journal of Robotics Research}, 38\penalty0
  (2-3):\penalty0 126--145, 2019.

\bibitem[Kulis \& Jordan(2011)Kulis and Jordan]{kulis2011revisiting}
Kulis, B. and Jordan, M.~I.
\newblock Revisiting k-means: New algorithms via bayesian nonparametrics.
\newblock \emph{arXiv preprint arXiv:1111.0352}, 2011.

\bibitem[Maddison et~al.(2016)Maddison, Mnih, and Teh]{maddison2016concrete}
Maddison, C.~J., Mnih, A., and Teh, Y.~W.
\newblock The concrete distribution: A continuous relaxation of discrete random
  variables.
\newblock \emph{arXiv preprint arXiv:1611.00712}, 2016.

\bibitem[Mnih et~al.(2015)Mnih, Kavukcuoglu, Silver, Rusu, Veness, Bellemare,
  Graves, Riedmiller, Fidjeland, Ostrovski, et~al.]{mnih2015human}
Mnih, V., Kavukcuoglu, K., Silver, D., Rusu, A.~A., Veness, J., Bellemare,
  M.~G., Graves, A., Riedmiller, M., Fidjeland, A.~K., Ostrovski, G., et~al.
\newblock Human-level control through deep reinforcement learning.
\newblock \emph{nature}, 518\penalty0 (7540):\penalty0 529--533, 2015.

\bibitem[Murali et~al.(2016)Murali, Garg, Krishnan, Pokorny, Abbeel, Darrell,
  and Goldberg]{murali2016tsc}
Murali, A., Garg, A., Krishnan, S., Pokorny, F.~T., Abbeel, P., Darrell, T.,
  and Goldberg, K.
\newblock Tsc-dl: Unsupervised trajectory segmentation of multi-modal surgical
  demonstrations with deep learning.
\newblock In \emph{2016 IEEE International Conference on Robotics and
  Automation (ICRA)}, pp.\  4150--4157. IEEE, 2016.

\bibitem[Nalisnick \& Smyth(2016)Nalisnick and Smyth]{nalisnick2016stick}
Nalisnick, E. and Smyth, P.
\newblock Stick-breaking variational autoencoders.
\newblock \emph{arXiv preprint arXiv:1605.06197}, 2016.

\bibitem[Neal(2000)]{neal2000markov}
Neal, R.~M.
\newblock Markov chain sampling methods for dirichlet process mixture models.
\newblock \emph{Journal of computational and graphical statistics}, 9\penalty0
  (2):\penalty0 249--265, 2000.

\bibitem[Niekum et~al.(2013)Niekum, Chitta, Barto, Marthi, and
  Osentoski]{niekum2013incremental}
Niekum, S., Chitta, S., Barto, A.~G., Marthi, B., and Osentoski, S.
\newblock Incremental semantically grounded learning from demonstration.
\newblock In \emph{Robotics: Science and Systems}, volume~9, pp.\  10--15607.
  Berlin, Germany, 2013.

\bibitem[Oliphant(2007)]{oliphant2007python}
Oliphant, T.~E.
\newblock Python for scientific computing.
\newblock \emph{Computing in science \& engineering}, 9\penalty0 (3):\penalty0
  10--20, 2007.

\bibitem[Paszke et~al.(2019)Paszke, Gross, Massa, Lerer, Bradbury, Chanan,
  Killeen, Lin, Gimelshein, Antiga, et~al.]{paszke2019pytorch}
Paszke, A., Gross, S., Massa, F., Lerer, A., Bradbury, J., Chanan, G., Killeen,
  T., Lin, Z., Gimelshein, N., Antiga, L., et~al.
\newblock Pytorch: An imperative style, high-performance deep learning library.
\newblock \emph{Advances in neural information processing systems},
  32:\penalty0 8026--8037, 2019.

\bibitem[Paulus et~al.(2020)Paulus, Maddison, and Krause]{paulus2020rao}
Paulus, M.~B., Maddison, C.~J., and Krause, A.
\newblock Rao-blackwellizing the straight-through gumbel-softmax gradient
  estimator.
\newblock \emph{arXiv preprint arXiv:2010.04838}, 2020.

\bibitem[Potapczynski et~al.(2020)Potapczynski, Loaiza-Ganem, and
  Cunningham]{potapczynski2020invertible}
Potapczynski, A., Loaiza-Ganem, G., and Cunningham, J.~P.
\newblock Invertible gaussian reparameterization: Revisiting the
  gumbel-softmax.
\newblock \emph{Advances in Neural Information Processing Systems}, 33, 2020.

\bibitem[Raffin et~al.(2021)Raffin, Hill, Gleave, Kanervisto, Ernestus, and
  Dormann]{stable-baselines3}
Raffin, A., Hill, A., Gleave, A., Kanervisto, A., Ernestus, M., and Dormann, N.
\newblock Stable-baselines3: Reliable reinforcement learning implementations.
\newblock \emph{Journal of Machine Learning Research}, 22\penalty0
  (268):\penalty0 1--8, 2021.
\newblock URL \url{http://jmlr.org/papers/v22/20-1364.html}.

\bibitem[Rezende et~al.(2014)Rezende, Mohamed, and
  Wierstra]{rezende2014stochastic}
Rezende, D.~J., Mohamed, S., and Wierstra, D.
\newblock Stochastic backpropagation and approximate inference in deep
  generative models.
\newblock In \emph{International conference on machine learning}, pp.\
  1278--1286. PMLR, 2014.

\bibitem[Roeder et~al.(2017)Roeder, Wu, and Duvenaud]{roeder2017sticking}
Roeder, G., Wu, Y., and Duvenaud, D.~K.
\newblock Sticking the landing: Simple, lower-variance gradient estimators for
  variational inference.
\newblock \emph{Advances in Neural Information Processing Systems},
  30:\penalty0 6925--6934, 2017.

\bibitem[Ross et~al.(2011)Ross, Gordon, and Bagnell]{ross2011reduction}
Ross, S., Gordon, G., and Bagnell, D.
\newblock A reduction of imitation learning and structured prediction to
  no-regret online learning.
\newblock In \emph{Proceedings of the fourteenth international conference on
  artificial intelligence and statistics}, pp.\  627--635. JMLR Workshop and
  Conference Proceedings, 2011.

\bibitem[Roychowdhury et~al.(2013)Roychowdhury, Jiang, and
  Kulis]{roychowdhury2013small}
Roychowdhury, A., Jiang, K., and Kulis, B.
\newblock Small-variance asymptotics for hidden markov models.
\newblock In \emph{Advances in Neural Information Processing Systems}, pp.\
  2103--2111, 2013.

\bibitem[Schaal et~al.(1997)]{schaal1997learning}
Schaal, S. et~al.
\newblock Learning from demonstration.
\newblock \emph{Advances in neural information processing systems}, pp.\
  1040--1046, 1997.

\bibitem[Schulman et~al.(2017)Schulman, Wolski, Dhariwal, Radford, and
  Klimov]{schulman2017proximal}
Schulman, J., Wolski, F., Dhariwal, P., Radford, A., and Klimov, O.
\newblock Proximal policy optimization algorithms.
\newblock \emph{arXiv preprint arXiv:1707.06347}, 2017.

\bibitem[Sethuraman(1994)]{sethuraman1994constructive}
Sethuraman, J.
\newblock A constructive definition of dirichlet priors.
\newblock \emph{Statistica sinica}, pp.\  639--650, 1994.

\bibitem[Shankar \& Gupta(2020)Shankar and Gupta]{shankar2020learning}
Shankar, T. and Gupta, A.
\newblock Learning robot skills with temporal variational inference.
\newblock In \emph{International Conference on Machine Learning}, pp.\
  8624--8633. PMLR, 2020.

\bibitem[Sharma et~al.(2018)Sharma, Sharma, Rhinehart, and
  Kitani]{sharma2018directed}
Sharma, A., Sharma, M., Rhinehart, N., and Kitani, K.~M.
\newblock Directed-info gail: Learning hierarchical policies from unsegmented
  demonstrations using directed information.
\newblock \emph{arXiv preprint arXiv:1810.01266}, 2018.

\bibitem[Sharma et~al.(2019)Sharma, Gu, Levine, Kumar, and
  Hausman]{sharma2019dynamics}
Sharma, A., Gu, S., Levine, S., Kumar, V., and Hausman, K.
\newblock Dynamics-aware unsupervised discovery of skills.
\newblock \emph{arXiv preprint arXiv:1907.01657}, 2019.

\bibitem[Shiarlis et~al.(2018)Shiarlis, Wulfmeier, Salter, Whiteson, and
  Posner]{shiarlis2018taco}
Shiarlis, K., Wulfmeier, M., Salter, S., Whiteson, S., and Posner, I.
\newblock {TACO}: Learning task decomposition via temporal alignment for
  control.
\newblock In Dy, J. and Krause, A. (eds.), \emph{Proceedings of the 35th
  International Conference on Machine Learning}, volume~80 of \emph{Proceedings
  of Machine Learning Research}, pp.\  4654--4663. PMLR, 10--15 Jul 2018.
\newblock URL \url{https://proceedings.mlr.press/v80/shiarlis18a.html}.

\bibitem[Stirn et~al.(2019)Stirn, Jebara, and Knowles]{stirn2019new}
Stirn, A., Jebara, T., and Knowles, D.
\newblock A new distribution on the simplex with auto-encoding applications.
\newblock \emph{Advances in Neural Information Processing Systems},
  32:\penalty0 13670--13680, 2019.

\bibitem[Such et~al.(2019)Such, Madhavan, Liu, Wang, Castro, Li, Zhi, Schubert,
  Bellemare, Clune, et~al.]{such2019atari}
Such, F.~P., Madhavan, V., Liu, R., Wang, R., Castro, P.~S., Li, Y., Zhi, J.,
  Schubert, L., Bellemare, M.~G., Clune, J., et~al.
\newblock An atari model zoo for analyzing, visualizing, and comparing deep
  reinforcement learning agents.
\newblock \emph{Proceedings of IJCAI 2019}, 2019.
\newblock URL \url{https://github.com/uber-research/atari-model-zoo}.

\bibitem[Sutton et~al.(1999)Sutton, Precup, and Singh]{sutton1999between}
Sutton, R.~S., Precup, D., and Singh, S.
\newblock Between mdps and semi-mdps: A framework for temporal abstraction in
  reinforcement learning.
\newblock \emph{Artificial intelligence}, 112\penalty0 (1-2):\penalty0
  181--211, 1999.

\bibitem[Teh et~al.(2006)Teh, Jordan, Beal, and Blei]{teh2006hierarchical}
Teh, Y.~W., Jordan, M.~I., Beal, M.~J., and Blei, D.~M.
\newblock Hierarchical dirichlet processes.
\newblock \emph{Journal of the american statistical association}, 101\penalty0
  (476):\penalty0 1566--1581, 2006.

\bibitem[vanRossum(1995)]{vanrossum1995python}
vanRossum, G.
\newblock Python reference manual.
\newblock \emph{Department of Computer Science [CS]}, \penalty0 (R 9525), 1995.

\bibitem[Vezhnevets et~al.(2017)Vezhnevets, Osindero, Schaul, Heess, Jaderberg,
  Silver, and Kavukcuoglu]{vezhnevets2017feudal}
Vezhnevets, A.~S., Osindero, S., Schaul, T., Heess, N., Jaderberg, M., Silver,
  D., and Kavukcuoglu, K.
\newblock Feudal networks for hierarchical reinforcement learning.
\newblock In \emph{International Conference on Machine Learning}, pp.\
  3540--3549. PMLR, 2017.

\bibitem[Williams(1992)]{williams1992simple}
Williams, R.~J.
\newblock Simple statistical gradient-following algorithms for connectionist
  reinforcement learning.
\newblock \emph{Machine learning}, 8\penalty0 (3):\penalty0 229--256, 1992.

\end{thebibliography}
\bibliographystyle{icml2022}

\newpage
\appendix
\onecolumn

\section{Relaxations} \label{app:relaxation}

We choose the following relaxations:
\begin{align}
    &\delta_{b_0^{(i)}=1} \approx b_0^{(i)}\\
    &\delta_{h_t^{(i)}=h_{t-1}^{(i)}} \approx 1 - ||h_t^{(i)} - h_{t-1}^{(i)}||_1 / 2\\
    &\eta(h_t^{(i)}) \approx \eta^\top h_t^{(i)}\\
    &p_\theta(b_t^{(i)}, h_t^{(i)}|h_{t-1}^{(i)}, s_t^{(i)}) \approx b_t^{(i)}\psi_{h_{t-1}^{(i)}}(s_t^{(i)})\eta^\top h_t^{(i)} + (1-b_t^{(i)})(1-\psi_{h_{t-1}^{(i)}}(s_t^{(i)}))\cdot (1 - ||h_t^{(i)} - h_{t-1}^{(i)}||_1 / 2)
\end{align}
Note that these relaxations $(1)$ match the objective they are relaxing when $b$'s are binary and $h$'s are one-hot; and $(2)$ that they are not unique: there are many sensible choices that could be used.

\section{Learning in Augmented Environments} \label{app:augmented}

In the augmented environment, taking an action corresponding to one of the skills means that the agent may actually interact with the environment for several timesteps. Na\"ively, one might think that simply using the usual Bellman backup update considering these enhanced actions as actions would be sufficient.
However, this neglects the fact that the reward credited to the enhanced action arises from multiple environment interactions, and that the discounting of the $Q$-value of the next state depends on the length of the enhanced action.
Recall that the usual Bellman equation is:
\begin{align}
    Q(s, a) &= \mathbb{E}_\pi\left[\displaystyle \sum_{t=1}^\infty \gamma^{t-1} R_{t}|S_0=s, A_0=a \right] \\
    &= \mathbb{E}_\pi\left[R_1 + \gamma Q(S_{1}, A_1)|S_0=s, A_0=a\right]
\end{align}
where the first term in the last equality ($R_1$) corresponds to the reward from a single timestep, and the second term ($\gamma Q(S_1, A_1)$) to the discounted return of all the following states. \\

For an augmented environment, let action $a$ result in $\tau$ interactions with the (unaugmented) environment.
Note that $\tau$ is a random variable, since skill terminations are stochastic, though of course $\tau = 1$ deterministically if $a$ is a primitive action.
We then have:
\begin{equation}
    Q(s, a) = \mathbb{E}_{\pi, \tau}\left[\displaystyle \sum_{t=1}^\tau \gamma^{t-1} R_t + \gamma^{\tau} Q(S_{\tau}, A_\tau)|S_0=s, A_0=a\right]
\end{equation}
As before, $t$ indexes interactions with the unaugmented environment during execution of a single enhanced action.
The first term ($\sum_{t=1}^\tau \gamma^{t-1}R_t$) corresponds to the reward directly credited to the (possibly enhanced) action, and the second term ($\gamma^\tau Q(S_\tau, A_\tau)$) is the discounted expected return from future (possibly enhanced) actions. As a sanity check, note that in the case where $\tau$ is always $1$, both equations match.

This modified Bellman equation implies that we must keep a slightly modified replay buffer.
The typical buffer $\{ (s_t, a_t, s_{t+1}, r_t) \}_t$, storing every interaction with the unaugmented environment, is insufficient.
We would be unable to determine when enhanced actions were taken or when they terminated, and therefore unable to credit them accordingly.
Further, such a buffer is wasteful, assuming that enhanced actions are common. Instead we should keep a buffer
$\{ (s_t, a_t, s_{t+\tau_t}, \sum_{t^{\prime}=t}^{t+\tau_t} \gamma^{t^{\prime} - t} r_{t^{\prime}}, \tau_t) \}$.
Here $a_t$ denotes the action taken in the augmented action space at time $t$ (possibly a primitive action, possibly an enhanced action).
$\tau_t$ is the (random) duration of that action.
We must keep $\tau_t$ to be able to discount $Q$ on the right hand side of the Bellman equation appropriately.
We need only put entries into the buffer at timesteps at which we take an action in the augmented environment. We do not need to record every interaction with the unaugmented environment.


\section{Experimental Details}\label{app:exp_details}

\paragraph{Proof-of-concept environment} For this experiment, we generate $1000$ expert trajectories using a manually designed policy. We then train our model for $500$ epochs. The options' sub-policies and termination functions consist of MLPs with two hidden layers of 16 units separated by a ReLU activation and followed by a Softmax activation. The parameters of all layers except the last one are shared across options. For the encoder, we use an LSTM layer with 32 hidden units, and MLPs with two hidden layers of 32 units and ReLU activations for both heads, with weights shared except for the last layer.
We use a learning rate of 0.005 with the Adam optimizer \citep{kingma2014adam} and a batch size of 128. The GS temperature parameter is initialized at 1 and annealed by a factor of 0.995 each epoch. $\lambda_{ent}$ is initialized at 5 and also annealed by a factor of 0.995 each epoch.
We check the option usage every 10 epochs ($n_K = 10^4$) and use $\delta = 0.5$ for our rule of increasing $K$.

\paragraph{Atari environments} For these experiments, we use $1000$ expert trajectories of length $300$ and train for $1500$ epochs. We use the RAM state of the Atari environments, where each state is a 128 bytes array, that we unpack into a 1024 bit array. The options' sub-policies and termination functions consist of a single linear layer with a Softmax activation per option. This choice is motivated by the fact that we do not want a single sub-policy to be able to fully reconstruct the expert trajectories. We use the same encoder as for the proof-of-concept environment, and the same optimizer with a learning rate of 0.001. The GS temperature is now annealed by a factor of 0.999. We keep the same values for $n_K$ and $\delta$.
We use similar architectures for sub-policies and termination in both CompILE and DDO, as well as for the encoder in CompILE. We fix the number of segments in CompILE to 7 for all runs.

To learn in the augmented environment, we use the PPO agent implemented by \citet{stable-baselines3}, with the default 'CnnPolicy' that takes as input the image state of the environment with the same preprocessing as done by  \citet{mnih2015human}. We use the implementation default parameters except for the 'n\_steps' variable (the number of environment steps used per update) that we set to 512. We also modify the replay buffer used during training to take into account the specific aspects of learning in an augmented environment mentioned in Appendix \ref{app:augmented}. This agent is trained for $3\cdot10^6$ (augmented) steps and the mean reward across the last $5\cdot10^5$ steps is used in Fig. \ref{fig:atari_exp_results}. For CompILE, we automatically terminate each enhanced action after 15 time steps for all augmented environments, as doing so was preferable to following the Poisson-sampled termination.

\section{Additional Experiments}\label{app:extra_exp}

\subsection{Proof-of-Concept Environment}

\begin{figure}[t]
    \centering
    \begin{subfigure}[b]{0.49\textwidth}
        \centering
         \includegraphics[width=\textwidth]{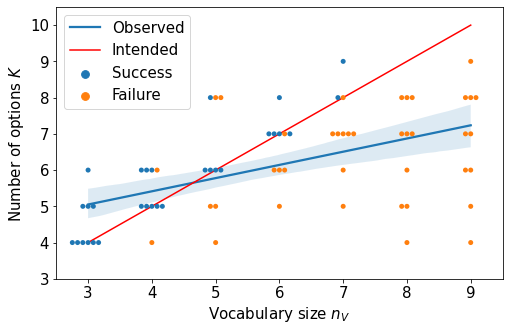}
    \end{subfigure}
    \begin{subfigure}[b]{0.49\textwidth}
        \centering
         \includegraphics[width=\textwidth]{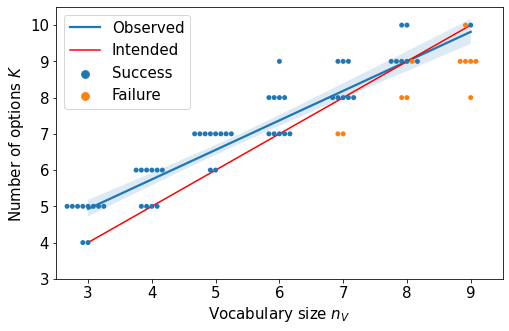}
    \end{subfigure}
    \caption{Results on our proof-of-concept environment, without entropy regularization (left panel), and with fixed entropy regularization (right panel).}
    \label{app_fig:ablations_1}
\end{figure}

We show in Figure \ref{app_fig:ablations_1} results analogous to those of Figure \ref{fig:toy_env_results}, except we do not use the entropy regularizer from Section \ref{sec:entropy_reg} (left panel), or we simply do not anneal it (right panel). We can see that, as mentioned in the main manuscript, not using the regularizer significantly degrades performance, although not using annealing (and keeping the regularizing coefficient fixed throughout training) does not have much of an impact.

\subsection{Atari Environments}

\begin{figure*}[h!]
    \centering
    \begin{subfigure}[b]{0.45\textwidth}
        \centering
         \includegraphics[width=\textwidth]{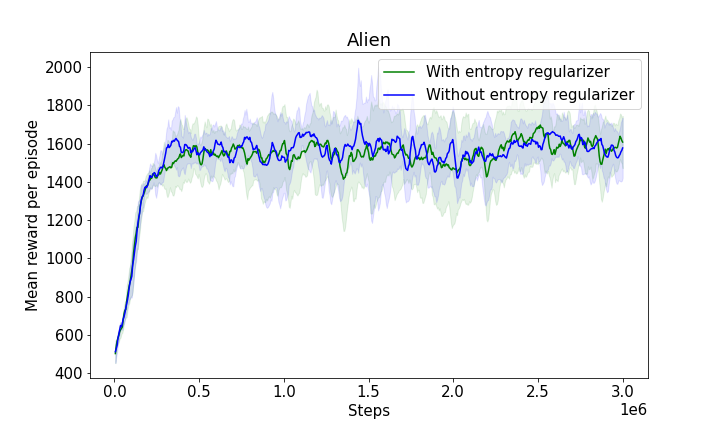}
    \end{subfigure}
    \begin{subfigure}[b]{0.45\textwidth}
        \centering
         \includegraphics[width=\textwidth]{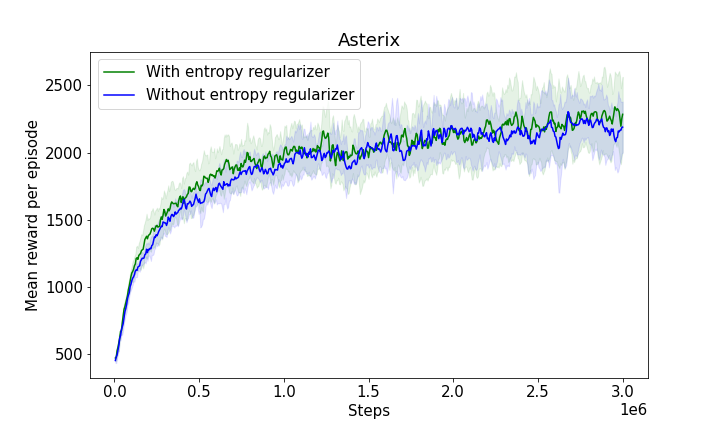}
    \end{subfigure}
    \begin{subfigure}[b]{0.45\textwidth}
        \centering
         \includegraphics[width=\textwidth]{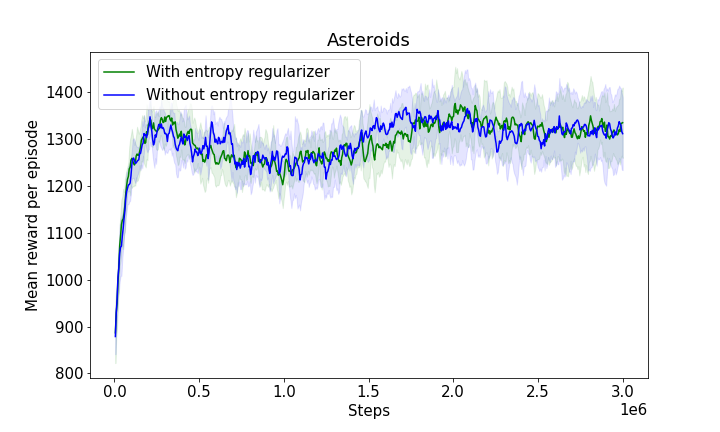}
    \end{subfigure}
    \begin{subfigure}[b]{0.45\textwidth}
        \centering
         \includegraphics[width=\textwidth]{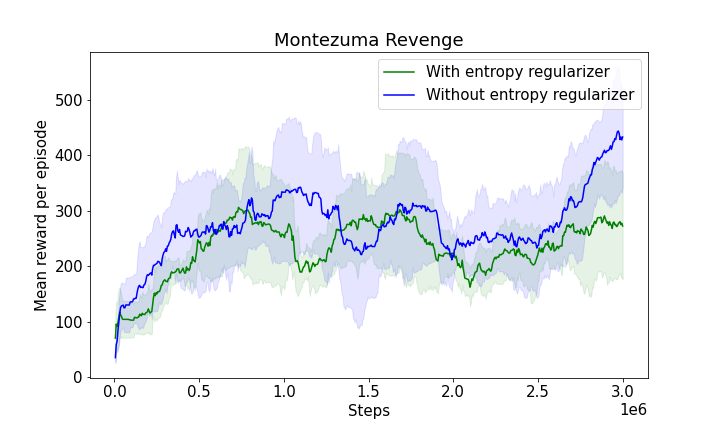}
    \end{subfigure}
    \begin{subfigure}[b]{0.45\textwidth}
        \centering
         \includegraphics[width=\textwidth]{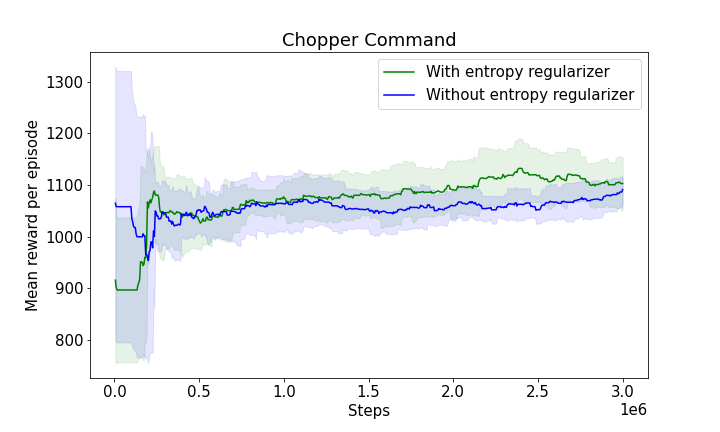}
    \end{subfigure}
    \begin{subfigure}[b]{0.45\textwidth}
        \centering
         \includegraphics[width=\textwidth]{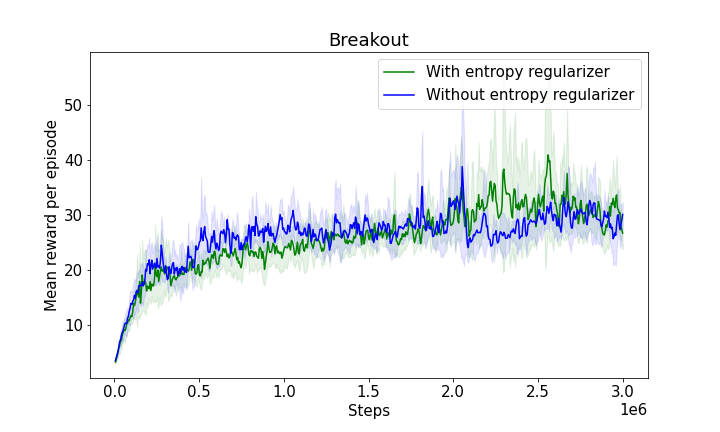}
    \end{subfigure}
    \begin{subfigure}[b]{0.45\textwidth}
        \centering
         \includegraphics[width=\textwidth]{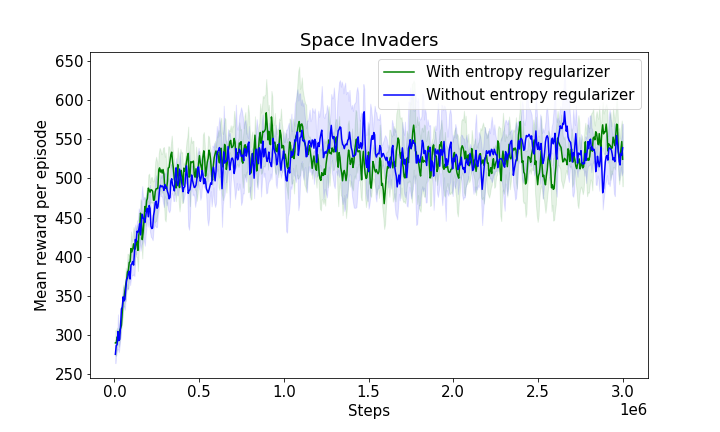}
    \end{subfigure}
    \caption{Evolution of the mean reward per episode during training with and without the entropy regularizer. Results are averaged across 5 runs.}
    \label{fig:entropy_comparison_atari}
\end{figure*}

As mentioned in the main manuscript, we ablate some of our choices. Figure \ref{fig:entropy_comparison_atari} shows our ablation results of using the entropy regularizer, green curves show results with the regularizer, and blue ones without. Across environments, performance is comparable, so that the entropy regularizer is not truly needed here. We highlight this result is opposite to what we found in our proof-of-concept environment, where the regularizer was fundamental to getting our method to work. Since adding the regularizer does not hurt performance in Atari environments and greatly helps in our proof-of-concept environment, we nonetheless recommend to use it as a default.

Tables \ref{tab:montezuma_traj_length}, \ref{tab:breakout_traj_length}, and \ref{tab:spaceInvaders_traj_length} show the results of ablations where the length of expert trajectories is changed for a subset of the Atari environments that we considered. We highlight that we did not cherry pick these environments, and the fact that we do not show analogous results for the missing environments was merely a matter of computational costs. While results do change significantly by varying trajectory length, both for our methods and the baselines, we can see that: $(1)$ our fixed-$K$ method consistently outperforms or remains competitive with both CompILE and DDO, the only exception being Montezuma's revenge with trajectories of length $50$, and $(2)$ our nonparametric method outperforms or remains competitive with nonparametric CompILE across all settings. We also highlight that DDO ran out of memory when using trajectories of length $1000$. We can thus see that our empirical superiority shown in the main manuscript was not due to a lucky choice of expert trajectory length.

\begin{table}[h!]
    \centering
    \caption{Reward on Montezuma's revenge, varying the length of expert trajectories. Columns requiring $K$, i.e. ``Ours'',``CompILE'', and ``DDO'', use $K=7$.}
    \vspace{1em}
    \begin{tabular}{r|c|c|c|c|c}
      Length & Ours & CompILE & DDO & Ours-NP & CompILE-NP \\
      \midrule
50 & 53.8 $\pm$ 7.6 & 113.1 $\pm$ 65.6 & 41.8 $\pm$ 20.9 & 274.0 $\pm$ 51.6 & 149.9 $\pm$ 73.7 \\
150 & 350.9 $\pm$ 44.8 & 0.3 $\pm$ 0.3 & 24.3 $\pm$ 19.2 & 344.9 $\pm$ 44.6 & 102.7 $\pm$ 91.7 \\
300 & 317.4 $\pm$ 48.6 & 0.0 $\pm$ 0.0 & 0.3 $\pm$ 0.2 & 261.0 $\pm$ 38.8 & 92.6 $\pm$ 80.0 \\
500 & 350.9 $\pm$ 42.5 & 0.0 $\pm$ 0.0 & 27.0 $\pm$ 0.0 & 391.2 $\pm$ 31.8 & 55.1 $\pm$ 48.4 \\
1000 & 328.3 $\pm$ 86.2 & 0.0 $\pm$ 0.0 & NA & 610.9 $\pm$ 74.4 & 66.6 $\pm$ 45.0 \\
    \end{tabular}
    \label{tab:montezuma_traj_length}
\end{table}

\begin{table}[h!]
    \centering
    \caption{Reward on Breakout, varying the length of expert trajectories. Columns requiring $K$ use $K=7$ for ``Ours'' and ``CompILE'' and $K=11$ for ``DDO''.}
    \vspace{1em}
    \begin{tabular}{r|c|c|c|c|c}
      Length & Ours & CompILE & DDO & Ours-NP & CompILE-NP \\
        \midrule
50 & 32.8 $\pm$ 1.0 & 16.0 $\pm$ 2.9 & 27.4 $\pm$ 1.3 & 31.6 $\pm$ 1.5 & 25.9 $\pm$ 2.6 \\
150 & 27.1 $\pm$ 1.3 & 20.0 $\pm$ 1.2 & 25.3 $\pm$ 1.9 & 22.9 $\pm$ 1.1 & 21.5 $\pm$ 1.1 \\
300 & 36.6 $\pm$ 3.0 & 18.6 $\pm$ 2.1 & 26.5 $\pm$ 1.7 & 31.6 $\pm$ 2.9 & 27.4 $\pm$ 2.9 \\
500 & 28.0 $\pm$ 2.0 & 18.4 $\pm$ 1.6 & 20.6 $\pm$ 0.0 & 22.6 $\pm$ 1.1 & 24.0 $\pm$ 3.4 \\
1000 & 23.0 $\pm$ 1.7 & 16.4 $\pm$ 3.2 & NA & 19.8 $\pm$ 1.1 & 17.9 $\pm$ 0.9 \\
    \end{tabular}
    \label{tab:breakout_traj_length}
\end{table}

\begin{table}[h!]
    \centering
    \caption{Reward on Space Invaders, varying the length of expert trajectories. Columns requiring $K$, i.e. ``Ours'',``CompILE'', and ``DDO'', use $K=7$.}
    \vspace{1em}
    \begin{tabular}{r|c|c|c|c|c}
      Length & Ours & CompILE & DDO & Ours-NP & CompILE-NP \\
        \midrule
50 & 488.1 $\pm$ 10.5 & 367.7 $\pm$ 18.1 & 414.1 $\pm$ 15.6 & 418.2 $\pm$ 26.6 & 440.2 $\pm$ 17.8 \\
150 & 562.0 $\pm$ 21.6 & 401.2 $\pm$ 14.8 & 475.0 $\pm$ 17.3 & 492.0 $\pm$ 16.1 & 447.3 $\pm$ 20.3 \\
300 & 583.0 $\pm$ 16.2 & 438.8 $\pm$ 45.6 & 479.3 $\pm$ 19.9 & 531.0 $\pm$ 17.7 & 469.9 $\pm$ 15.9 \\
500 & 562.9 $\pm$ 18.9 & 432.6 $\pm$ 29.2 & 468.8 $\pm$ 27.1 & 478.4 $\pm$ 12.9 & 469.2 $\pm$ 17.2 \\
1000 & 552.5 $\pm$ 19.4 & 420.8 $\pm$ 24.4 & NA & 490.8 $\pm$ 16.3 & 502.3 $\pm$ 25.5 \\
    \end{tabular}
    \label{tab:spaceInvaders_traj_length}
\end{table}


\end{document}